\pdfoutput=1

\documentclass[11pt]{article}

\usepackage{acl}

\usepackage{times}
\usepackage{latexsym}

\usepackage{pifont}
\usepackage{soul}

\usepackage[T1]{fontenc}

\usepackage[utf8]{inputenc}

\usepackage{microtype}

\usepackage{inconsolata}

\usepackage{graphicx}
\usepackage{algorithm}
\usepackage{algorithmic}
\usepackage{amsfonts}
\usepackage{multirow}
\usepackage{makecell}
\usepackage{amsmath}
\usepackage{amsthm}
\usepackage{color}
\usepackage{subcaption} 
\usepackage{soul}
\usepackage{booktabs}   
\usepackage{graphicx}   
\usepackage{multirow}   
\usepackage{makecell}   
\usepackage{colortbl}   
\usepackage{amsmath}    

\definecolor{green}{RGB}{0, 160, 70} 
\definecolor{scientificred}{RGB}{192, 0, 0} 
\usepackage{xcolor}
\usepackage{tcolorbox}

\usepackage{tikz}
\definecolor{yellowFillColor}{HTML}{FFF2CC}
\definecolor{yellowStrokeColor}{HTML}{D6B656}

\definecolor{myGreenFill}{HTML}{D5E8D4}
\definecolor{myGreenStroke}{HTML}{82B366}

\definecolor{myPinkFill}{HTML}{F8CECC}
\definecolor{myPinkStroke}{HTML}{B85450}

\title{SynGraph: A Dynamic Graph-LLM Synthesis Framework for Sparse Streaming User Sentiment Modeling}

\author
{
Xin Zhang$^{1}$, Qiyu Wei$^{1}$, Yingjie Zhu$^{2}$, Linhai Zhang$^{3}$, Deyu Zhou$^{4}$, Sophia Ananiadou$^{1}$\\
$^{1}$The University of Manchester \quad
$^{2}$Harbin Institute of Technology \quad \\
$^{3}$King’s College London \quad
$^{4}$Southeast University\\
\texttt{\{xin.zhang-41, qiyu.wei\}@postgrad.manchester.ac.uk}\\
\texttt{sophia.ananiadou@manchester.ac.uk, zhuyj@stu.hit.edu.cn}\\
\texttt{linhai.zhang@kcl.ac.uk, d.zhou@seu.edu.cn}
}

\begin{document}
\maketitle
\begin{abstract}
User reviews on e-commerce platforms exhibit dynamic sentiment patterns driven by temporal and contextual factors. Traditional sentiment analysis methods focus on static reviews, failing to capture the evolving temporal relationship between user sentiment rating and textual content. Sentiment analysis on streaming reviews addresses this limitation by modeling and predicting the temporal evolution of user sentiments. However, it suffers from data sparsity, manifesting in temporal, spatial, and combined forms.
In this paper, we introduce \textsc{SynGraph}, a novel framework designed to address data sparsity in sentiment analysis on streaming reviews. \textsc{SynGraph} alleviates data sparsity by categorizing users into mid-tail, long-tail, and extreme scenarios and incorporating LLM-augmented enhancements within a dynamic graph-based structure.
Experiments on real-world datasets demonstrate its effectiveness in addressing sparsity and improving sentiment modeling in streaming reviews.

\end{abstract}

\section{Introduction}

\begin{figure}[!t] 
  \centering
  \includegraphics[width=0.5\textwidth]{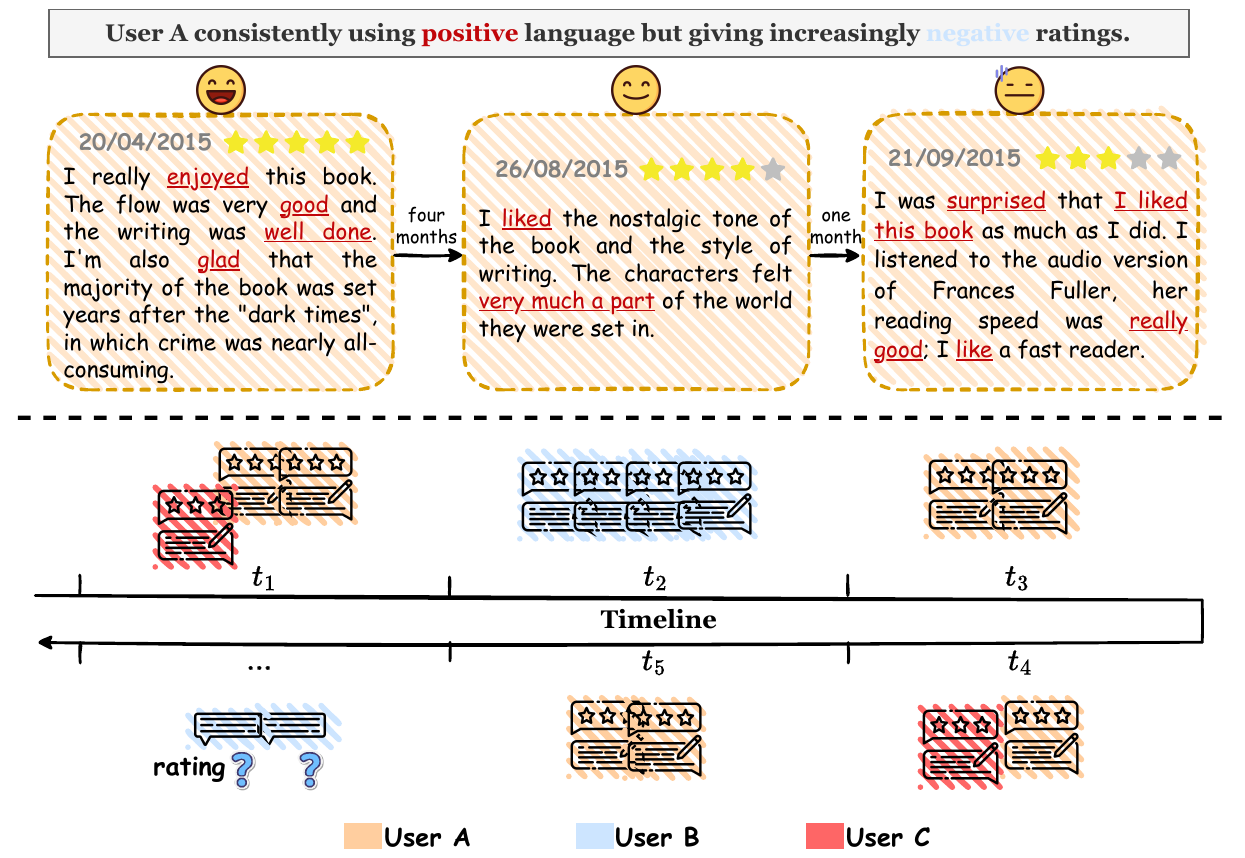}
    \caption{Sentiment Evolution in Streaming Reviews.}
    \label{fig:intro}
\end{figure}

User reviews on e-commerce platforms provide valuable insights into customer preferences, product performance, and user satisfaction. These reviews reflect users' momentary sentiments and exhibit temporal dynamics as users interact with products over time. Traditional sentiment analysis methods, however, predominantly operate in static contexts, where individual reviews are treated as independent data points \cite{DBLP:conf/acl/TangQL15,DBLP:conf/emnlp/ChenSTLL16,DBLP:conf/ACMicec/KimS07,DBLP:conf/ic3i/MittalAR22,DBLP:journals/kbs/ZhangZZ25}. This paradigm overlooks the inherent temporal dependencies and evolving nature of user sentiments, which are crucial for capturing the full spectrum of user behavior in real-world scenarios. 
For instance, as shown in Figure~\ref{fig:intro}, a user’s sentiment rating may initially be highly positive (e.g., ``I really \underline{enjoyed} this book'') but gradually shift to neutral (e.g., ``I was \underline{surprised} that I liked this book as much as I did'') in subsequent interactions. These transitions highlight the dynamic relationship between sentiment changes and textual cues, require models to learn sentiment change patterns from historical reviews, and leverage this understanding to predict future sentiment ratings.

As a solution to these limitations, the task of ``Sentiment Analysis on Streaming Reviews'' has been proposed, aiming to model the temporal evolution of user sentiments across sequential time windows and predict user future sentiments \cite{DBLP:conf/emnlp/ZhangZZ23,DBLP:conf/emnlp/WuSS023}. This task emphasizes the interplay between numerical sentiment scores and their associated textual content over time. 
However, this task also introduces significant challenges, primarily due to the problem of data sparsity \cite{DBLP:conf/ijcai/Guo13,DBLP:conf/sigir/DuY00022,DBLP:journals/corr/abs-2403-06139}.
Previous studies have commonly introduced graph structure information to supplement sparse user data in static scenarios (\citealp{DBLP:journals/corr/abs-2302-02151,DBLP:conf/sigir/WangZSWW23,DBLP:conf/sigir/ChenWHHXLH022}). Despite this, in streaming scenarios, the inclusion of graph structure information further complicates the classification and discussion of sparse users \cite{10387583}.
As shown in Figure~\ref{fig:intro}, some users leave reviews at irregular and widely spaced intervals, making it challenging for temporal models to capture sequential dependencies (User A: temporal sparsity). Others have limited social connections, resulting in a lack of neighbor information to support sentiment analysis (User B: spatial sparsity). Additionally, some users exhibit both temporal and spatial sparsity, further complicating predictive modeling (User C: combined sparsity). Existing methods fail to effectively address these sparsity issues, struggling to model temporal dependencies and spatial interactions.

To tackle these challenges, we propose \textsc{SynGraph}, a novel framework tailored for Sentiment Analysis on Streaming Reviews. 
\textsc{SynGraph} operates within a dynamic graph structure and leverages LLM-augmented data synthesis to enhance sparse user representations. Specifically, \textsc{SynGraph} employs a decompose-and-recompose strategy to categorize users into three scenarios—mid-tail, long-tail, and extreme scenarios—based on their temporal and spatial characteristics. For each scenario, \textsc{SynGraph} integrates three core components:
(a)~\emph{Local and global graph understanding}: Captures both micro-level (local) and macro-level (global) patterns of user-product interactions.
(b)~\emph{High-order relation understanding}: Explores second-order and higher-order dependencies to enrich graph representations.
(c)~\emph{LLM-augmented data synthesis}: Utilizes large language models to induce supplementary data for sparse users, enhancing representation robustness.
Combining these components, \textsc{SynGraph} effectively addresses the sparsity challenges in streaming reviews and provides a robust solution for learning dynamic sentiment patterns. 
Our main contributions can be summarized as follows:
\begin{itemize}
    \item We introduce \textsc{SynGraph}, a novel framework that integrates dynamic graph modeling with LLM-augmented data synthesis to address the problem of data sparsity in sentiment analysis on streaming reviews.
    \item We propose a decompose-and-recompose strategy to categorize users into different sparsity levels (mid-tail, long-tail, and extreme) and apply tailored profile-enhancing techniques to synthesize data for each category.
    \item We conduct extensive experiments on real-world datasets, demonstrating that \textsc{SynGraph} significantly improves sentiment prediction accuracy while effectively mitigating the impact of temporal and spatial sparsity.
\end{itemize}

\section{Problem Formulation}

\subsection{Continuous-Time Dynamic Graph}
To model evolving user-product interactions, we represent the system as a Continuous-Time Dynamic Graph (CTDG; \citet{DBLP:journals/corr/abs-2203-10480}). Formally, for a time span \(T_{\text{span}} = [t_0, t_n]\), the dynamic graph is defined as 
\(
    G_{T_{\text{span}}} = \bigl(G_{t_0}, O_{[t_1, t_n]}\bigr),
\)
where \(G_{t_0}\) is the initial graph at \(t_0\) and \(O_{[t_1, t_n]}\) is a sequence of timestamped events. Each event \(o_i \in O_{[t_1, t_n]}\) represents either a node update \((v_i, t_i)\) (adding a new user or product) or an edge update \((u_i, p_i, t_i)\) (a user–product interaction).

\subsection{Streaming User Review Graph}

\begin{figure}[!t] 
    \centering
    \includegraphics[width=\linewidth]{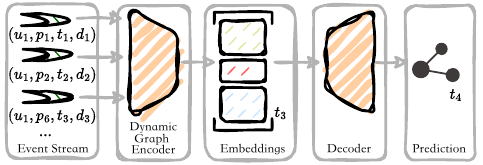}
    \caption{Streaming User Review Graph.}
    \label{fig:model}
\end{figure}
In the streaming review scenario, we observe a temporal sequence of reviews
\(
    E = \{\mathcal{E}_1, \mathcal{E}_2, \dots, \mathcal{E}_T\},
\)
with each review \(\mathcal{E}_i = (u_i, p_i, t_i, d_i, y_i)\) capturing the user \(u_i\), product \(p_i\), timestamp \(t_i\), review content \(d_i\), and sentiment \(y_i\). We instantiate the CTDG framework to form our Streaming User Review Graph \(G_T\), where users and products are nodes and each review creates a directed edge \((u_i, p_i, t_i)\) annotated with \(\{d_i, y_i\}\). In this graph, new reviews trigger node and edge updates, seamlessly mirroring the CTDG structure.
As illustrated in Figure~\ref{fig:model}, interactions arrive as a continuous stream and are processed by a dynamic graph encoder that generates temporal embeddings reflecting evolving user preferences. These embeddings feed into a decoder that predicts future user sentiment at time \(t_4\):
\(
    y = f\Bigl(\mathcal{E}_t \,\Big|\, \{\mathcal{E}_1, \dots, \mathcal{E}_{t-1}\}\Bigr).
\)
By leveraging the CTDG structure, our approach naturally captures both temporal dependencies and structural evolution in user-product interactions, thereby improving sentiment prediction.

\section{\textsc{SynGraph} Framework}

\begin{figure*}[!t]
\centering
\includegraphics[width=1\textwidth]{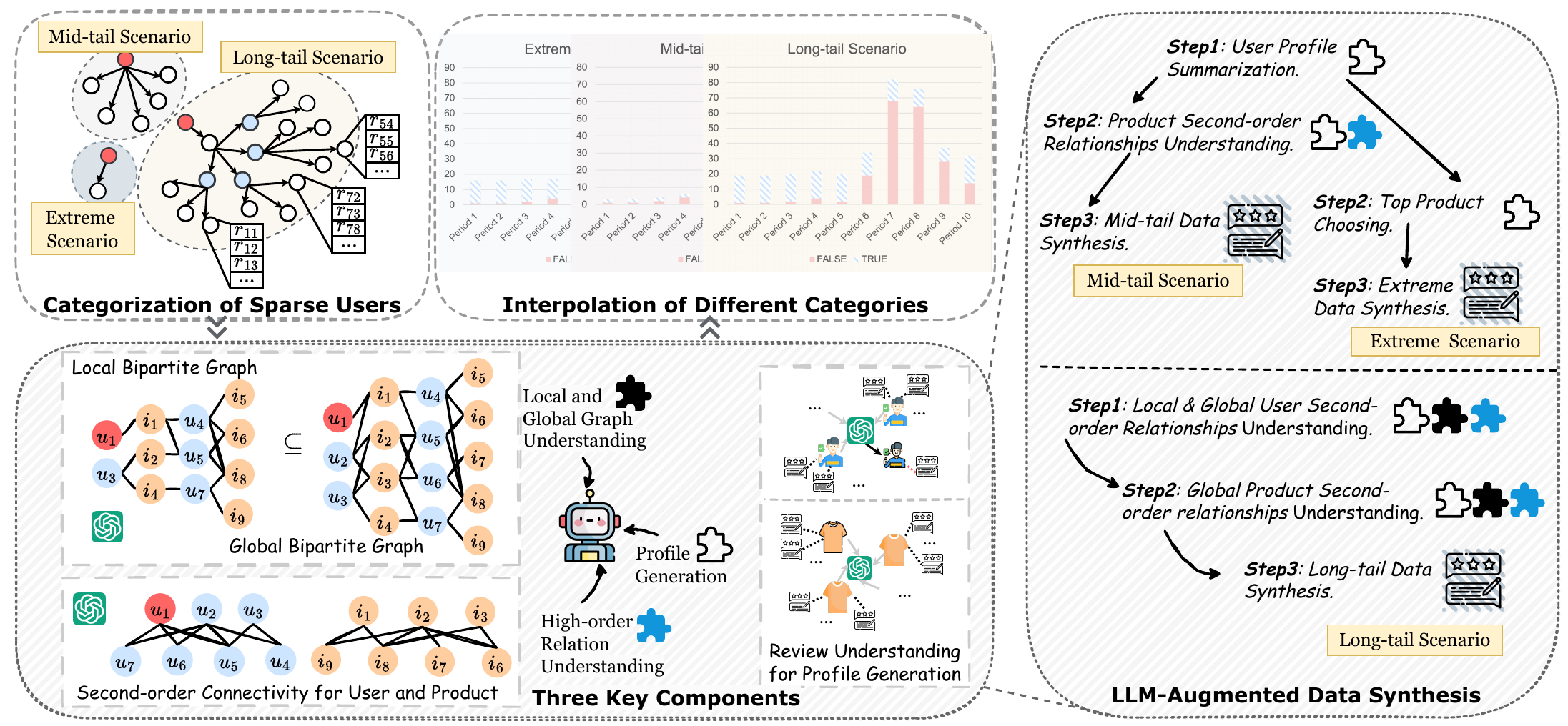}
\caption{Overview of \textsc{SynGraph}, a three-stage framework for addressing data sparsity in sentiment analysis on streaming reviews. The three stages include: (1) Categorization of sparse users—identifying user types based on temporal and structural patterns; (2) LLM-augmented data Synthesis—enhancing user and product profiles using first- and second-order relationships; (3) Interpolation of different categories—integrating synthesized data to improve representation learning and sentiment prediction.}
\label{fig:Framework}
\end{figure*}

We propose \textsc{SynGraph}, a three-stage framework that addresses data sparsity in Sentiment Analysis on Streaming Reviews by leveraging LLM-augmented dynamic graphs. The \textsc{SynGraph} framework consists of three key stages: (1) Categorization of sparse users—identifying user types based on their temporal and structural patterns; (2) LLM-augmented representation learning—enhancing user and product profiles using first- and second-order relationships; (3) Interpolation of different categories—integrating synthesized data to improve representation learning and sentiment prediction.


\subsection{Categorization of Sparse Users}
To better model user behaviors in streaming sentiment analysis, we classify users into three sparse categories:

\noindent\emph{(\romannumeral 1) Mid-tail Users:}  
These users actively contribute reviews within specific time frames but exhibit fluctuations in engagement across different intervals. To quantify this variability, we measure the variance in user interactions over a set of time intervals \(T\):
\(
\sigma^2_{I(u_m)} = \frac{1}{T} \sum_{t=1}^{T} \left( I_t(u_m) - \bar{I}(u_m) \right)^2,
\)
where \( I_t(u_m) \) represents the number of interactions by user \( u_m \) in interval \( t \), and \( \bar{I}(u_m) \) is the mean interaction count across all intervals. A higher variance indicates greater fluctuations, suggesting that the user's engagement pattern is inconsistent over time.  
\noindent\emph{(\romannumeral 2) Long-tail Users:}  
Long-tail users contribute a limited number of reviews and have sparse first-order interactions, meaning they have few direct neighbors in the interaction graph. This limits the effectiveness of traditional neighbor-based analysis. However, these users often maintain a substantial number of second-order connections, allowing for indirect information propagation. We define first-order connectivity as:
\(
    C_1(u_l) = \sum_{v \in \mathcal{N}_1(u_l)} w(u_l, v),
\)
where \( \mathcal{N}_1(u_l) \) is the set of direct neighbors of user \( u_l \), and \( w(u_l, v) \) denotes the interaction weight between \( u_l \) and neighbor \( v \). A small \( C_1(u_l) \) value indicates that user \( u_l \) has sparse immediate connections, whereas a high \( C_2(u_l) \) value suggests that second-order neighbors provide rich contextual signals.
\noindent\emph{(\romannumeral 3) Extreme Users:}  
These users are highly sparse both in terms of review activity and connectivity within the interaction graph, making them challenging to model with traditional approaches. Their lack of both first- and second-order connections results in minimal information propagation. We define the overall connectivity score as:
\(
    C(u_e) = \sum_{v \in \mathcal{N}(u_e)} w(u_e, v),
\)
where \( \mathcal{N}(u_e) \) denotes the set of all neighbors connected to \( u_e \). A near-zero \( C(u_e) \) indicates that the user is almost completely isolated in the network.  

As illustrated in the top-left part of Figure~\ref{fig:Framework} (``Categorization of Sparse Users''), these three categories of sparse users exhibit distinct distributions. In the figure, the three types of sparse users are highlighted in red, the first-order neighbors (products) in white, and the second-order homogeneous neighbors (users) in blue, with each node containing a subset of relevant reviews.

\subsection{LLM-Augmented Data Synthesis}
In this section, we introduce an LLM-augmented Data Synthesis strategy, using LLM to synthesize user and product representations while incorporating structural neighbor information from dynamic graphs. This stage consists of two main parts: Key Components for LLM-augmented Data Synthesis and LLM-augmented Data Synthesis.

\noindent\textbf{Key Components for LLM-Augmented Data Synthesis:} To effectively handle data sparsity in streaming sentiment analysis, we introduce three essential components that enable LLMs to synthesize high-quality augmented data.

\noindent\emph{(1) Local-Global Graph Understanding:} 
In our framework, the user profile is enhanced by supplementing it with contextual information derived from both local and global neighbor data. Specifically, the local information is obtained from the immediate neighbors that reflect recent and direct interactions, while the global information is collected from a broader set of neighbors, capturing long-term trends and overall network context. Our approach employs LLM to interpret and enrich the user's profile based on these local and global cues. 

\noindent\emph{(2) High-order Relation Understanding:} 
Instead of relying solely on direct connections, we extract second-order relationships to improve user-product profiles. Given a user \( u \) with neighbors \( \mathcal{N}(u) \), we define their second-order neighborhood as:
\(
    \mathcal{N}^2(u) = \left(\bigcup_{v \in \mathcal{N}(u)} \mathcal{N}(v)\right) \setminus \{u\}.
\)

\noindent\emph{(3) Profile Generation:} 
For sparse users and products lacking explicit profiles, we utilize LLMs to synthesize synthetic profile based on existing reviews. These profiles can be directly extracted from user-generated content or inferred through second-order neighborhood information within the interaction graph.
Formally, we define the profile generation function as:
\(    
\mathcal{P}(x) = \text{LLM}\big( \mathcal{R}(x), P_x \big),
\)
where \(\mathcal{R}(x)\) is the selected review set for entity \(x\) (which can be a user or a product), and \(P_x\) is the corresponding guiding prompt. In our implementation, we use:\(P_u\) for generating user profiles, \(P_{\text{prod}}\) for generating product profiles, and \(P_{\text{data}}\) for synthesizing the final review data.

\noindent\textbf{LLM-Augmented Data Synthesis:} After obtaining the three key components, we combine these three methods and use them in three sparse categories.

\noindent\emph{(a) Mid-tail Data Synthesis:} 
Mid-tail users have moderate interactions, allowing profile generation from historical reviews and second-order product relationships. 
Formally, the profile of a mid-tail user $u_m$ is generated as:
\(
    \mathcal{P}(u_m) = \text{LLM} \big( \mathcal{R}(u_m), P_{u} \big).
\)
To incorporate second-order product relationships, we refine the product profile set as:
\(
    \mathcal{P}_{\text{set}}(u_m) = \text{LLM} \big( P_{\text{prod}}, \mathcal{N}^2(\mathcal{N}(u_m)) \big),
\)
where $P_{\text{prod}}$ is the prompt used for generating product profiles.
The final synthesized data is then constructed by integrating both user and product profiles:
\begin{equation}
    \mathcal{D}_M = \text{LLM} \big( P_{\text{data}}, \mathcal{P}(u_m), \mathcal{P}_{\text{set}}(u_m) \big).
\end{equation}

\noindent\emph{(b) Long-tail Data Synthesis:} 
Long-tail users have significantly fewer interactions, necessitating the use of neighbor-based augmentation. Their profiles are constructed using both local and global second-order relationships:
\(
    \mathcal{P}(u_l) = \text{LLM} \big( \mathcal{N}^2_{\text{local}}(u_l), \mathcal{N}^2_{\text{global}}(u_l), \mathcal{R}_{\text{chosen}}(u_l), P_u \big),
\)
where $\mathcal{N}^2_{\text{local}}(u_l)$ and $\mathcal{N}^2_{\text{global}}(u_l)$ denote the second-order neighbors from the local and global graphs, respectively.
Similarly, product profiles for long-tail users are defined as:
\(
    \mathcal{P}_{\text{set}}(u_l) = \text{LLM} \big( P_{\text{prod}}, \mathcal{N}^2(\mathcal{N}(u_l)) \big).
\)
The synthesized data for long-tail users is obtained by:
\begin{equation}
    \mathcal{D}_L = \text{LLM} \big( P_{\text{data}}, \mathcal{P}(u_l), \mathcal{P}_{\text{set}}(u_l) \big).
\end{equation}

\noindent\emph{(c) Extreme Data Synthesis:} 
For extreme users with minimal or no interactions, we introduce pseudo-data synthesis based on high-rated products. The user profile is synthesized as:
\(
    \mathcal{P}(u_e) = \text{Profile}(u_e),
\)
where \(\text{Profile}(u_e)\) denotes a pre-defined or externally obtained profile for the extreme user \(u_e\).
Popular products are incorporated as proxies:
\(
    \mathcal{P}_{\text{set}}(u_e) = \text{LLM}\Bigl( \mathcal{R}_{\text{chosen}}(P_{\text{high}}), P_{\text{prod}} \Bigr),
\)
where \(P_{\text{high}}\) denotes a selected set of high-rated products and \(\mathcal{R}_{\text{chosen}}(P_{\text{high}})\) is the corresponding review set.
The final synthesized data for extreme users is given by:
\begin{equation}
    \mathcal{D}_E = \text{LLM} \big( P_{\text{data}}, \mathcal{P}(u_e), \mathcal{P}_{\text{set}}(u_e) \big).
\end{equation}

By leveraging LLM-based profile generation, second-order relationship modeling, and local-global graph understanding, \textsc{SynGraph} effectively synthesizes data for sparse users, enhancing model stability across varying sparsity levels.

\begin{table*}[!t]
\centering
\renewcommand{\arraystretch}{1.2}
\resizebox{1.98\columnwidth}{!}{ 
\begin{tabular}{lccccccccccccc}
\toprule
\multirow{2}{*}{\textbf{Category}} & \multicolumn{2}{c}{\textbf{Normal}} & \multicolumn{2}{c}{\textbf{Mid-tail}} & \multicolumn{2}{c}{\textbf{Long-tail}} & \multicolumn{2}{c}{\textbf{Extreme}} & \hspace{1.2cm} & \multicolumn{3}{c}{\textbf{Interpolated Review Count}} \\ 
\cmidrule(lr){2-3} \cmidrule(lr){4-5} \cmidrule(lr){6-7} \cmidrule(lr){8-9} \cmidrule(lr){11-13}
 & Count & Proportion & Count & Proportion & Count & Proportion & Count & Proportion & & Mid-tail & Long-tail & Extreme \\ 
\midrule
Magazine\_Subscriptions & 183 & (52.59\%) & 8 & (2.30\%) & 154 & (44.25\%) & 3 & (0.86\%) & & 67 & 1287 & 26 \\ 
Appliances & 11 & (23.40\%) & 2 & (4.26\%) & 19 & (40.36\%) & 15 & (31.91\%) & & 15 & 158 & 126 \\ 
Gift\_Cards & 203 & (44.42\%) & 45 & (9.85\%) & 209 & (45.73\%) & 0 & (0.00\%) & & 358 & 1753 & 0 \\ 
\bottomrule
\end{tabular}}
\vspace{-0.1in}
\caption{\textbf{Left:} User distribution across different sparsity levels.  \textbf{Right:} Number of interpolated reviews.}
\label{tab:user_sparsity_interpolation}
\end{table*}

\subsection{Interpolation of Different Categories}
After generating enriched representations for each user category, we apply data interpolation across different user groups to improve sentiment modeling and ensure data balance.
To quantify the interpolation need for a given user $u$, we define the interpolation factor as:
\(
    I_f(u) = \frac{1}{T} \sum_{t=1}^{T} \mathbf{1}_{\{ I_t(u) = 0 \}},
\)
where $\mathbf{1}_{\{ I_t(u) = 0 \}}$ is an indicator function that returns 1 if the user $u$ has no interactions at time step $t$, ensuring that interpolation is applied where necessary.
Using our interpolation position search method, we identify the missing data points for each category and apply interpolation accordingly. The total number of interpolated interactions for a given category $C$ is computed as:
\begin{equation}
    I_{\text{total}}(C) = \sum_{u \in C} \sum_{t=1}^{T} \mathbf{1}_{\{ I_t(u) = 0 \}},
\end{equation}
where $C$ represents a user category (mid-tail, long-tail, or extreme).
The interpolated data is then incorporated into the training pipeline, ensuring that each user maintains a minimum data availability threshold of 10 interactions.

\section{Experiments}
\begin{table*}[!t]
\centering
\renewcommand{\arraystretch}{1.2}
\resizebox{1.9\columnwidth}{!}{
\begin{tabular}{ccccccccc}

\toprule
\multirow{2}{*}{\textbf{Method}}  & \multicolumn{7}{c}{\textbf{Evaluation Metrics}} & \multirow{2}{*}{\textit{RMSE Reduction}}  \\ 
\cmidrule{2-8}
& Accuracy ($\uparrow$) & Precision ($\uparrow$) & Recall ($\uparrow$) & F1 ($\uparrow$) & MSE ($\downarrow$) & RMSE ($\downarrow$) & MAE ($\downarrow$) \\ 
\midrule

\multicolumn{9}{c}{\cellcolor{gray!15} \textbf{\textit{Dataset: Magazine\_Subscriptions}}} \\
BiLSTM+Att & 0.6910 & 0.4040 & 0.4054 & 0.4019 & 1.5021 & 1.2256 & 0.5837 & - \\
BiLSTM+Att$^{*}$ & \cellcolor{blue!10}0.6831 & \cellcolor{blue!10}0.5966 & \cellcolor{blue!10}0.3667 & \cellcolor{blue!10}0.3976 & \cellcolor{blue!10}1.3862 & \cellcolor{blue!10}1.1774 & \cellcolor{blue!10}0.5628 & \textcolor{green}{$\downarrow$}3.93\% \\
Bert-Sequence & 0.6953 & 0.2589 & 0.3049 & 0.2791 & 1.2918 & 1.1366 & 0.5451 & - \\
Bert-Sequence$^{*}$ & \cellcolor{blue!10}0.7035 & \cellcolor{blue!10}0.4479 & \cellcolor{blue!10}0.5278 & \cellcolor{blue!10}0.4792 & \cellcolor{blue!10}0.5160 & \cellcolor{blue!10}0.7183 & \cellcolor{blue!10}0.3654 & \textcolor{green}{$\downarrow$}{36.80\%} \\
NGSAM & - & - & - & - & 1.2867&1.1343 &0.7772 & - \\
NGSAM$^{*}$ & - & - & - & - & \cellcolor{blue!10}0.9433 & \cellcolor{blue!10}0.9712 & \cellcolor{blue!10}0.6077 &  \textcolor{green}{$\downarrow$}{14.38\%} \\
CHIM &0.6084 & 0.3549 & 0.3002 & 0.3073& 1.5392 & 1.2406& 0.6831 & -\\
CHIM$^{*}$ &\cellcolor{blue!10}0.6995 & \cellcolor{blue!10}0.2453 & \cellcolor{blue!10}0.2768 &\cellcolor{blue!10}0.2553 & \cellcolor{blue!10}1.8798 & \cellcolor{blue!10}1.3711& \cellcolor{blue!10}0.6557&  \textcolor{red}{$\uparrow$}{10.52\%}  \\
IUPC & 0.7039 & 0.2671 & 0.3499 & 0.3016 & 0.9442 & 0.9717 & 0.4635 & - \\
 IUPC$^{*}$ & \cellcolor{blue!10}0.7420& \cellcolor{blue!10}0.4800 & \cellcolor{blue!10}0.5283   & \cellcolor{blue!10}0.4979 & \cellcolor{blue!10}0.5721 & \cellcolor{blue!10}0.7564   & \cellcolor{blue!10}0.3349  &  \textcolor{green}{$\downarrow$}{22.16\%}  \\ 
\textbf{DC-DGNN} & 0.7554 & 0.4290 & 0.4107 & 0.4016 & 0.7768 & 0.8814 & 0.3820 & - \\
\textbf{DC-DGNN}$^{*}$ & \cellcolor{blue!10}\textbf{0.7983} & \cellcolor{blue!10}\textbf{0.6879} & \cellcolor{blue!10}\textbf{0.5853} & \cellcolor{blue!10}\textbf{0.5385} & \cellcolor{blue!10}\textbf{0.4206} & \cellcolor{blue!10}\textbf{0.6485} & \cellcolor{blue!10}\textbf{0.2575} & \textcolor{green}{$\downarrow$}{26.42\%} \\

\midrule
 \multicolumn{9}{c}{\cellcolor{gray!15} \textbf{\textit{Dataset: Appliances}}} \\

BiLSTM+Att & 0.5000 & 0.1250 & 0.2500 & 0.1667 & 1.3250 & 1.1511 & 0.7250 & - \\
BiLSTM+Att$^{*}$ & \cellcolor{blue!10}0.7250 & \cellcolor{blue!10}0.6230 & \cellcolor{blue!10}0.5417 & \cellcolor{blue!10}0.5605 & \cellcolor{blue!10}0.3500 & \cellcolor{blue!10}0.5916 & \cellcolor{blue!10}0.3000 & \textcolor{green}{$\downarrow$}{48.61\%} \\
Bert-Sequence & 0.7143 & 0.2381 & 0.3333 & 0.2778 & 1.0476 & 1.0235 & 0.4762 & - \\
Bert-Sequence$^{*}$ &\cellcolor{blue!10}0.6042 &\cellcolor{blue!10}0.2062 &  \cellcolor{blue!10}0.2404& \cellcolor{blue!10}0.2178 & \cellcolor{blue!10}0.9583 & \cellcolor{blue!10}0.9789  & \cellcolor{blue!10}0.5417  &  \textcolor{green}{$\downarrow$}{4.36\%}  \\
NGSAM & - & - & - & - & 0.7007&0.8371 & 0.6010  & -\\ 
NGSAM$^{*}$ & - & - & - & - &\cellcolor{blue!10}0.5069 & \cellcolor{blue!10}0.7120 & \cellcolor{blue!10}0.4763 &  \textcolor{green}{$\downarrow$}{14.94\%}  \\ 
CHIM & 0.5500 &0.2244& 0.2414&0.2230& 0.9500&0.9747 & 0.6000  & - \\ 
CHIM$^{*}$ & \cellcolor{blue!10}0.6000 & \cellcolor{blue!10}0.2468 &\cellcolor{blue!10}0.2614& \cellcolor{blue!10}0.2453 &\cellcolor{blue!10}0.5500& \cellcolor{blue!10}0.7416 & \cellcolor{blue!10}0.4500 &  \textcolor{green}{$\downarrow$}{23.92\%} \\ 
IUPC & 0.7143 & 0.2381 & 0.3333 & 0.2778 & 1.0476 & 1.0235 & 0.4762  & -\\
IUPC$^{*}$ & \cellcolor{blue!10}0.6042  & \cellcolor{blue!10}0.2510& \cellcolor{blue!10}0.3212  &\cellcolor{blue!10}0.2483& \cellcolor{blue!10}0.9583 & \cellcolor{blue!10}0.9789& \cellcolor{blue!10}0.5417   &  \textcolor{green}{$\downarrow$}{4.36\%}  \\ 
 \textbf{DC-DGNN} & 0.7143 & 0.2381 & 0.3333 & 0.2778 & 1.0476 & 1.0235 & 0.4762  & -\\
\textbf{DC-DGNN}$^{*}$ & \cellcolor{blue!10}\textbf{0.8571} & \cellcolor{blue!10}\textbf{0.5441} & \cellcolor{blue!10}\textbf{0.5833} & \cellcolor{blue!10}\textbf{0.5625} & \cellcolor{blue!10}\textbf{0.6667} & \cellcolor{blue!10}\textbf{0.8165} & \cellcolor{blue!10}\textbf{0.2857} & \textcolor{green}{$\downarrow$}{20.22\%} \\

\midrule
 \multicolumn{9}{c}{\cellcolor{gray!15} \textbf{\textit{Dataset: Gift\_Cards}}} \\
BiLSTM+Att & 0.7439 & 0.2690 & 0.2034 & 0.1778 & 0.5213 & 0.7220 & 0.3273 & - \\
BiLSTM+Att$^{*}$ & \cellcolor{blue!10}0.7749 & \cellcolor{blue!10}0.3066 & \cellcolor{blue!10}0.2631 & \cellcolor{blue!10}0.2727 & \cellcolor{blue!10}0.5201 & \cellcolor{blue!10}0.7211 & \cellcolor{blue!10}0.3079 & \textcolor{green}{$\downarrow$}{0.12\%} \\
Bert-Sequence & 0.8754 & 0.2189 & 0.2500 & 0.2334 & 0.3064 & 0.5535 & 0.1717  & -\\
Bert-Sequence$^{*}$ & \cellcolor{blue!10}0.8443 & \cellcolor{blue!10}0.4417 & \cellcolor{blue!10}0.4053 & \cellcolor{blue!10}0.4201 & \cellcolor{blue!10}0.2909 & \cellcolor{blue!10}0.5393 &\cellcolor{blue!10}0.1905&  \textcolor{green}{$\downarrow$}{2.57\%}\\
NGSAM & - & - & - & - &0.4152 & 0.6444 & 0.3463 & -\\
NGSAM$^{*}$ & - & - & - & - &\cellcolor{blue!10}0.2544 & \cellcolor{blue!10}0.5044 & \cellcolor{blue!10}0.3116 &  \textcolor{green}{$\downarrow$}{21.73\%}\\
CHIM & 0.8754 & 0.2189 & 0.2500 & 0.2334 & 0.3064 & 0.5535 & 0.1717 & - \\
 CHIM$^{*}$ & \cellcolor{blue!10}0.7671& \cellcolor{blue!10}0.3698 & \cellcolor{blue!10}0.3346 &\cellcolor{blue!10}0.3475 &\cellcolor{blue!10}0.6080 & \cellcolor{blue!10}0.7798 & \cellcolor{blue!10}0.3312 &  \textcolor{red}{$\uparrow$}{40.89\%}\\
 \textbf{DC-DGNN} & 0.8754 & 0.2189 & 0.2500 & 0.2334 & 0.3064 & 0.5535 & 0.1717  & -\\
\textbf{DC-DGNN}$^{*}$ & \cellcolor{blue!10}\textbf{0.8956} & \cellcolor{blue!10}\textbf{0.4384} & \cellcolor{blue!10}\textbf{0.5000} & \cellcolor{blue!10}\textbf{0.4671} & \cellcolor{blue!10}\textbf{0.1145} & \cellcolor{blue!10}\textbf{0.3383} & \cellcolor{blue!10}\textbf{0.1077} & \textcolor{green}{$\downarrow$}{38.88\%} \\

\bottomrule
\end{tabular}
}
\vspace{-0.05in}
\caption{Performance comparison of sentiment analysis models on streaming user reviews across three real-world Amazon datasets. $\downarrow$ indicates that lower values are better, while $\uparrow$ indicates that higher values are better. Models marked with $^{*}$ denote those trained on datasets augmented with the three types of synthesized data via the \textsc{SynGraph} framework, and these rows are highlighted in blue. \textit{RMSE Reduction} represents performance improvement, where \textcolor{green}{$\downarrow$} indicates an performance improvement after data augmentation, while \textcolor{red}{$\uparrow$} indicates a performance drop. }
\vspace{-0.2in}
\label{tab:main_result}
\end{table*}

\subsection{Experiments Setup}

\noindent\textbf{Dataset statistics.} We utilize three datasets from the Amazon dataset collection \cite{DBLP:conf/emnlp/NiLM19}, specifically Magazine\_Subscriptions, Appliances, and Gift\_Cards, chosen for their relatively smaller data sizes. To preserve the integrity of the original data distribution, we retain them in their raw form without additional preprocessing. Users in each dataset are categorized into mid-tail, long-tail, and extreme users based on the definitions established in this paper. (The detailed user categorization process is provided in the appendix.)
The distribution of users across different sparsity levels is presented in Table~\ref{tab:user_sparsity_interpolation}.

\noindent\textbf{Baselines.} We compare our approach against two categories of baseline models:  
(1)~Text-based models: BiLSTM+Att, BERT-Sequence \citep{DBLP:conf/naacl/DevlinCLT19}.  
(2)~User and product-based models: CHIM \cite{DBLP:conf/emnlp/Amplayo19}, IUPC \citep{DBLP:conf/coling/LyuFG20}, NGSAM \cite{DBLP:conf/aaai/ZhouZZH21}, and DC-DGNN \cite{DBLP:conf/emnlp/ZhangZZ23}.  
Among these, DC-DGNN is specifically designed for continuous dynamic graph learning on streaming data. Models marked with $^{*}$ denote those trained on datasets augmented with the three types of synthesized data synthesized via the \textsc{SynGraph} framework.

\noindent\textbf{Implementation details.} For user and product embeddings, all models utilize a $128$-dimensional representation. The batch size is set to $8$, and the learning rate is $3e-5$, with training conducted for $2$ epochs. Sentiment analysis is formulated as a classification problem, and model performance is evaluated using seven metrics:  Accuracy, Precision, Recall, F1-score, Mean Squared Error (MSE), Root Mean Squared Error (RMSE), and Mean Absolute Error (MAE).
The dataset is split into training and test sets at a $9:1$ ratio. Table~\ref{tab:user_sparsity_interpolation} also presents the statistics of interpolated review counts across different categories.

\begin{table*}[!t]
\centering
\renewcommand{\arraystretch}{1.2}
\resizebox{1.8\columnwidth}{!}{
\begin{tabular}{l ccccccc}
\toprule
\multirow{2}{*}{\textbf{Method}} & \multicolumn{7}{c}{\textbf{Evaluation Metrics}} \\ 
\cmidrule{2-8}
& Accuracy ($\uparrow$) & Precision ($\uparrow$) & Recall ($\uparrow$) & F1 ($\uparrow$) & MSE ($\downarrow$) & RMSE ($\downarrow$) & MAE ($\downarrow$) \\ 
\midrule

\rowcolor{gray!15} \multicolumn{8}{c}{\textbf{\textit{Dataset: Magazine\_Subscriptions}}} \\
DC-DGNN$^{*}$ & 0.7983 & 0.6879 & 0.5853 & 0.5385 & 0.4206 & 0.6485 & 0.2575 \\
DC-DGNN-M & \textbf{0.8155} & 0.6785 & \textbf{0.6044} & \textbf{0.6225} & 0.4292 & 0.6551 & 0.2489 \\
DC-DGNN-L & 0.7725 & \textbf{0.6826} & 0.5029 & 0.4661 & 0.5365 & 0.7324 & 0.2961 \\
DC-DGNN-E & 0.8112 & 0.5715 & 0.5521 & 0.5484 & \textbf{0.4077} & \textbf{0.6385} & \textbf{0.2446} \\

\midrule
\rowcolor{gray!15} \multicolumn{8}{c}{\textbf{\textit{Dataset: Appliances}}} \\
\textbf{DC-DGNN}$^{*}$ & \textbf{0.8571} & \textbf{0.5441} & \textbf{0.5833} & \textbf{0.5625} & \textbf{0.6667} & \textbf{0.8165} & \textbf{0.2857} \\
DC-DGNN-M & 0.7143 & 0.2381 & 0.3333 & 0.2778 & 1.0476 & 1.0235 & 0.4762 \\
DC-DGNN-L & 0.6667 & 0.2593 & 0.3111 & 0.2828 & 0.6190 & 0.7868 & 0.4286 \\
DC-DGNN-E & 0.7143 & 0.2381 & 0.3333 & 0.2778 & 1.0476 & 1.0235 & 0.4762 \\

\midrule
\rowcolor{gray!15} \multicolumn{8}{c}{\textbf{\textit{Dataset: Gift\_Cards}}} \\
\textbf{DC-DGNN}$^{*}$ & \textbf{0.8956} & \textbf{0.4384} & \textbf{0.5000} & \textbf{0.4671} & \textbf{0.1145} & \textbf{0.3383} & \textbf{0.1077} \\
DC-DGNN-M & 0.8754 & 0.2234 & 0.2500 & 0.2359 & 0.1448 & 0.3805 & 0.1313 \\
DC-DGNN-L & 0.8754 & 0.2189 & 0.2500 & 0.2334 & 0.3064 & 0.5535 & 0.1717 \\

\bottomrule
\end{tabular}
}
\vspace{-0.08in}
\caption{Results of interpolating sparse user data across different categories. {DC-DGNN}$^{*}$ represents the results obtained by inducing data from Mid-tail, Long-tail, and Extreme user categories. DC-DGNN-M refers to the results obtained by inducing data from only Mid-tail users, DC-DGNN-L from only Long-tail users, and DC-DGNN-E from only Extreme users.}
\label{tab:category_cmp}
\end{table*}

\subsection{Main Results}
The performance on Magazine\_Subscriptions, Appliances, and Gift\_Cards datasets, illustrating the impact of incorporating synthesized data into sentiment analysis of streaming user reviews, are presented in Table~\ref{tab:main_result}. The key observations are as follows:
(1) Across all three datasets, with a total of $18$ experimental configurations spanning six different models, we observe that $15$ out of $18$ settings exhibit a significant improvement in performance when leveraging the \textsc{SynGraph} framework. This result underscores the effectiveness of \textsc{SynGraph} in enhancing sentiment analysis in streaming settings.
(2) Among all evaluated models, DC-DGNN and its augmented variant DC-DGNN$^{*}$ consistently achieve the best performance, both before and after incorporating synthesized data. This finding highlights the critical role of dynamic user and product modeling in sentiment analysis for streaming reviews.
(3) A comparative analysis of test set results before and after incorporating synthesized data reveals a steady performance improvement after augmentation. This phenomenon suggests that the model demonstrates improved generalization and robustness, validating the ability of \textsc{SynGraph} to synthesize diverse and effective user data.

\subsection{Ablation Study}
\vspace{-0.01in}
To validate the effectiveness of each proposed component, we conducted ablation experiments on DC-DGNN to assess the efficiency of data synthesis for each category, denoted as -M, -L, and -E for mid-tail, long-tail, and extreme users, respectively.
As shown in Table~\ref{tab:category_cmp}, we found that combining data from all three categories generally resulted in the best performance across various datasets, such as Appliances and Gift\_Cards. In some cases, using only one type of supplementation led to the optimal outcome, as observed in Magazine\_Subscriptions.
This is because the datasets considered in this study are small-scale datasets, and introducing more data could introduce additional noise, potentially leading to a decrease in predictive performance. 
Similarly, there was no change in performance in the -M and -E cases of the Appliances dataset, likely due to the small number of synthesized data introduced. This is reasonable, as attempting to improve performance by introducing only a few data points, as shown in Table~\ref{tab:user_sparsity_interpolation}, is also unlikely. 
As for the -L case of the Gift\_Cards dataset, overfitting still occurred, likely due to the severe imbalance of the original labels in this dataset, with proportions corresponding to labels 5, 4, 3, 2, and 1 being [0.9258, 0.0519, 0.0111, 0.0074, 0.0037] respectively.
Introducing a large amount of similar data under the long-tail scenario exacerbated this imbalance. However, it is worth mentioning that the overfitting phenomenon during training on the Gift\_Cards dataset was mitigated to some extent when combining the synthesized data from all three categories.

\begin{figure}[!ht] 
  \centering
  \includegraphics[width=0.49\textwidth]{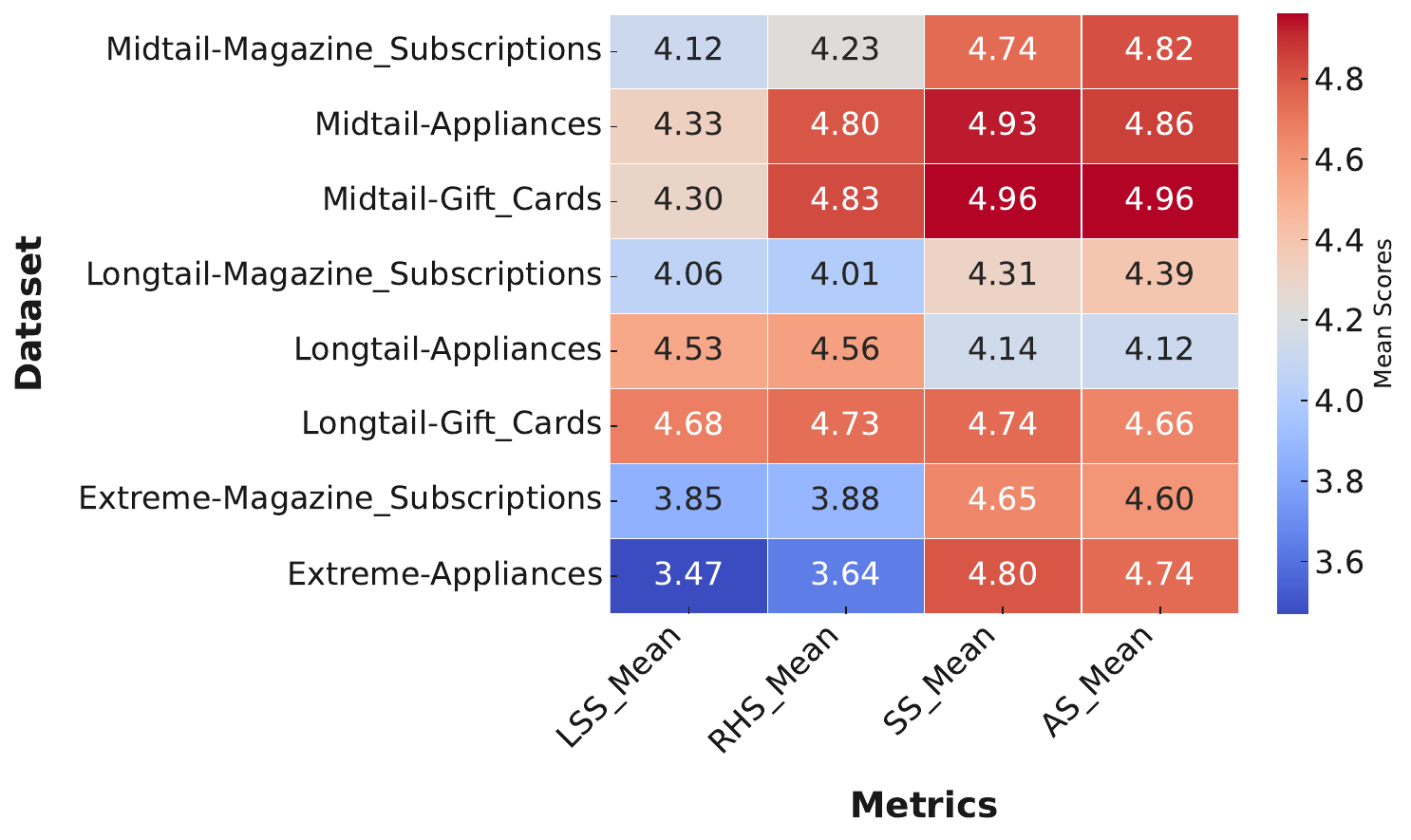}
  \caption{GPT-4 evaluation of synthesized reviews across different sparsity categories.}
  \label{fig:evaluation_results}
\end{figure}

\subsection{Evaluation of Synthesized Data}
\vspace{-0.05in}
To assess the quality of the LLM-synthesized data, we conducted an evaluation using the latest GPT-4\footnote{\url{https://openai.com/gpt-4}}  model across four key metrics:
(1)~Language Style Similarity (LSS): Measures the alignment between the synthesized review and the user's writing style.
(2)~Rating Habit Similarity (RHS): Evaluates the consistency between the synthesized rating and the user's historical rating patterns.
(3)~Sentiment Similarity (SS): Assesses whether the sentiment of the synthesized review aligns with product reviews.
(4)~Aspect Similarity (AS): Determines whether the synthesized review focuses on relevant product attributes.
Figure~\ref{fig:evaluation_results} presents the results across different sparsity categories and datasets. We observe that mid-tail users exhibit the highest consistency across all metrics, while extreme sparsity scenarios result in lower LSS and RHS scores due to limited user history. Even in sparse cases, SS and AS scores remain stable, indicating that the model effectively maintains product-level coherence.

\subsection{Vocabulary Richness Analysis} 
\begin{figure}[!t] 
  \centering
  \includegraphics[width=0.49\textwidth]{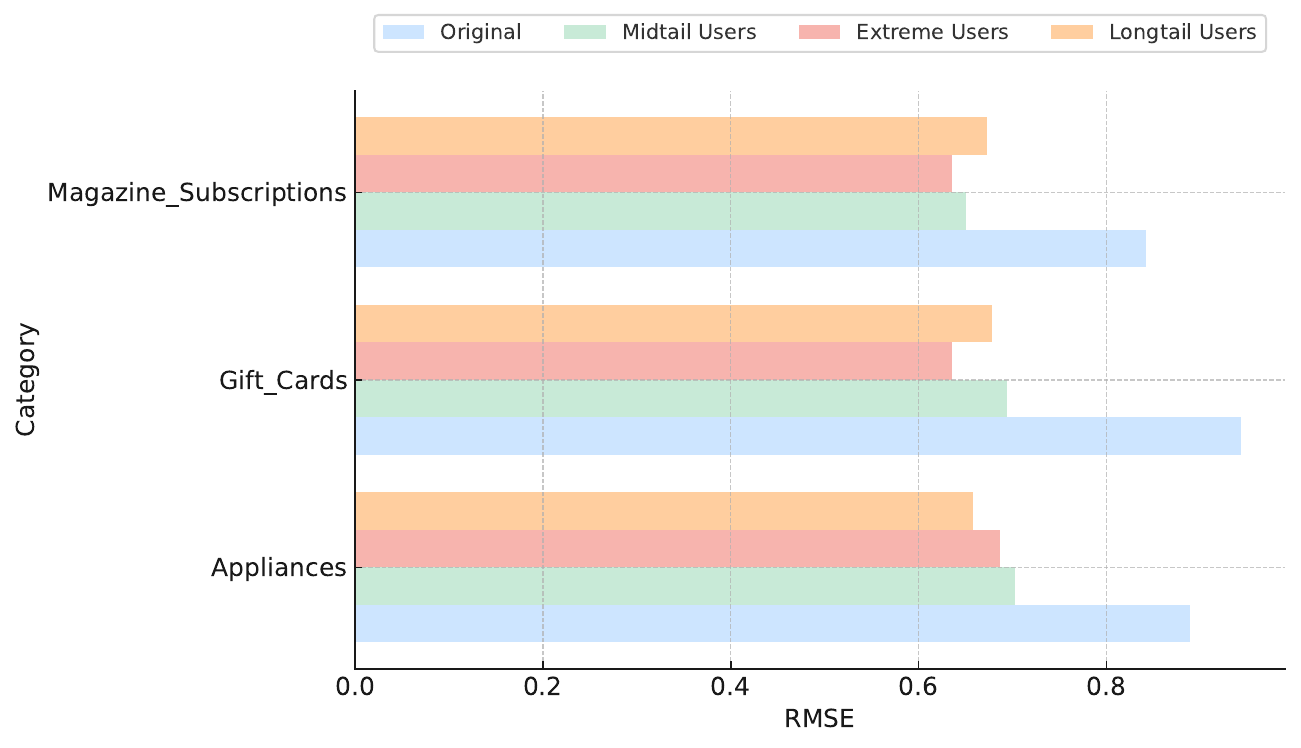}
    \caption{Comparison of vocabulary richness between LLM-synthesized and original data across different sparsity categories.}
    \label{fig:vocabulary_richness}
\end{figure}

To evaluate the quality of the LLM-Synthesized data, we employ NLTK\footnote{\url{https://www.nltk.org/}} to compute the overall average vocabulary richness across different sparsity categories. We then compare these values with those of the original dataset, as shown in Figure~\ref{fig:vocabulary_richness}.  
Our analysis reveals that the LLM-synthesized text exhibits a vocabulary richness pattern consistent with prior findings \cite{DBLP:conf/emnlp/LiZL023}, suggesting a potential limitation in lexical diversity. Specifically, across all sparsity categories and datasets, the vocabulary richness of the LLM-synthesized text is consistently lower than that of the original dataset. Furthermore, the richness levels remain relatively stable across different categories of synthesized data, indicating that while the LLM effectively synthesizes supplementary content, it may introduce a degree of lexical homogeneity.

\section{Related Work}
Data sparsity is a persistent challenge in e-commerce recommendation, where traditional methods enhance preference representations by leveraging user attributes \cite{DBLP:conf/nips/VolkovsYP17,DBLP:conf/sigir/ZhuSSC20}, integrating social networks \cite{DBLP:conf/recsys/SedhainSBXC14,DBLP:conf/sigir/DuY00022}, or transferring knowledge across domains \cite{DBLP:conf/cikm/HuZY18,DBLP:conf/wsdm/0008T20,DBLP:conf/sigir/GaoWLWLYZL023,DBLP:conf/sigir/ZhuGZXXZL021}. Few-shot learning techniques further mitigate sparsity by leveraging semantic associations in graph-structured data \cite{DBLP:conf/cikm/WangZWZLXG18,DBLP:journals/corr/abs-2302-02151,DBLP:conf/sigir/WangZSWW23,DBLP:conf/sigir/ChenWHHXLH022} or employing meta-learning to enhance adaptability across domains \cite{DBLP:conf/sigir/WuZ23,DBLP:conf/kdd/Lu0S20,DBLP:conf/kdd/LeeIJCC19}. However, these approaches treat sparsity as a static issue, failing to model the temporal evolution of user interactions and the structural complexity of dynamic streaming data.

Recent advances in LLMs offer new opportunities to address sparsity by leveraging their reasoning abilities and extensive pretrained knowledge. Prompt engineering has been explored for mathematical QA \cite{yu2024metamath}, chain-of-thought reasoning \cite{liang-etal-2023-prompting}, symbolic data generation \cite{ye-etal-2023-generating}, and hybrid human–LLM annotation strategies \cite{staab2024beyond,li-etal-2023-coannotating,li-etal-2023-robust}. In recommendation systems, LLMs have been employed to enrich sparse user data by integrating first-order neighbor information \cite{DBLP:journals/corr/abs-2311-00423} or inferring user profiles from textual descriptions \cite{DBLP:conf/emnlp/SunL0CAZJ23}. However, these methods often overlook the structural complexity of graphs, limiting their effectiveness in leveraging higher-order relationships. Moreover, temporal dependencies are frequently ignored, leading to suboptimal modeling of evolving user preferences in streaming environments.

Despite these advancements, an effective framework that jointly models temporal dynamics, high-order structural dependencies, and LLM-based augmentation remains underexplored. Our work fills this gap by introducing a unified approach that systematically integrates these factors, offering a more comprehensive solution to the challenges in sentiment analysis on streaming user reviews.

\section{Conclusion}
We propose \textsc{SynGraph}, a dynamic graph-based framework that mitigates data sparsity in sentiment analysis on streaming reviews. By integrating LLM-augmented enhancements, it effectively models local and global structures, high-order relationships, and supplementary data. It adapts to different sparsity scenarios through a flexible decomposition and recombination mechanism. Experiments on three real-world Amazon datasets demonstrate its effectiveness in improving sentiment analysis on streaming reviews under sparse conditions.

\section*{Acknowledgement}
We would like to thank the anonymous reviewers for their insightful comments and suggestions. This work was supported by the British Heart Foundation Manchester Research Excellence Award (RE/24/130017). We also acknowledge the CSF3 at the University of Manchester for providing GPU resources. Xin is  supported by the UoM-CSC Joint Scholarship.


\section*{Limitations}
While our data synthesis approach effectively mitigates user data sparsity and demonstrates strong performance, several aspects warrant further investigation:  

\begin{itemize}
    \item \textbf{Neighbor Selection Strategy:} To balance efficiency and computational cost, we adopt a random sampling strategy when selecting next-hop neighbors. While this method proves effective in our study, particularly for datasets with limited sample sizes, its impact may vary in large-scale scenarios where sampling variance becomes more pronounced. Future work could explore more principled selection mechanisms that optimize for both efficiency and representational quality.  

    \item \textbf{LLM Understanding of Graph Structures:} Our framework integrates local and global graph structures to enhance representation learning. However, we do not explicitly analyze how LLMs differentiate between these two types of graphs or the extent to which each contributes to the final representations. A deeper investigation into LLMs’ ability to process and leverage hierarchical graph structures could further improve the robustness of graph-based data augmentation.  

    \item \textbf{Mitigating Hallucination in Data Synthesis:} Like many applications of LLMs, our approach is susceptible to hallucination, where synthesized data may not always faithfully reflect real-world distributions. While our framework benefits from LLM-augmented synthesis, ensuring the reliability and factual consistency of synthesized data remains an open challenge. Future research could focus on refining prompt designs or incorporating external validation mechanisms to enhance the trustworthiness of synthesized content.  
\end{itemize}

Despite these considerations, our approach provides a solid foundation for addressing data sparsity in sentiment analysis on streaming reviews, and we believe that further refinements in these areas can further enhance its applicability.

\bibliography{SynGraph}

\clearpage

\appendix


\section{Experiment Details}
\label{sec:exp}
\noindent\textbf{Sparse User Categorization.}  
Users with more than five reviews are considered non-sparse, while those with five or fewer are categorized as sparse based on clustering results. Table~\ref{tab:dif_user_statistic} shows the proportion of reviews contributed by these two groups, which together constitute the majority of reviews. Thus, we focus on these user types, where users with 0 to 5 reviews are classified as sparse (long-tail or extreme), and those with 5 to 10 reviews are non-sparse.  

\begin{figure*}[htbp]
  \centering
  \begin{subfigure}[b]{0.41\textwidth}
    \includegraphics[width=\textwidth]{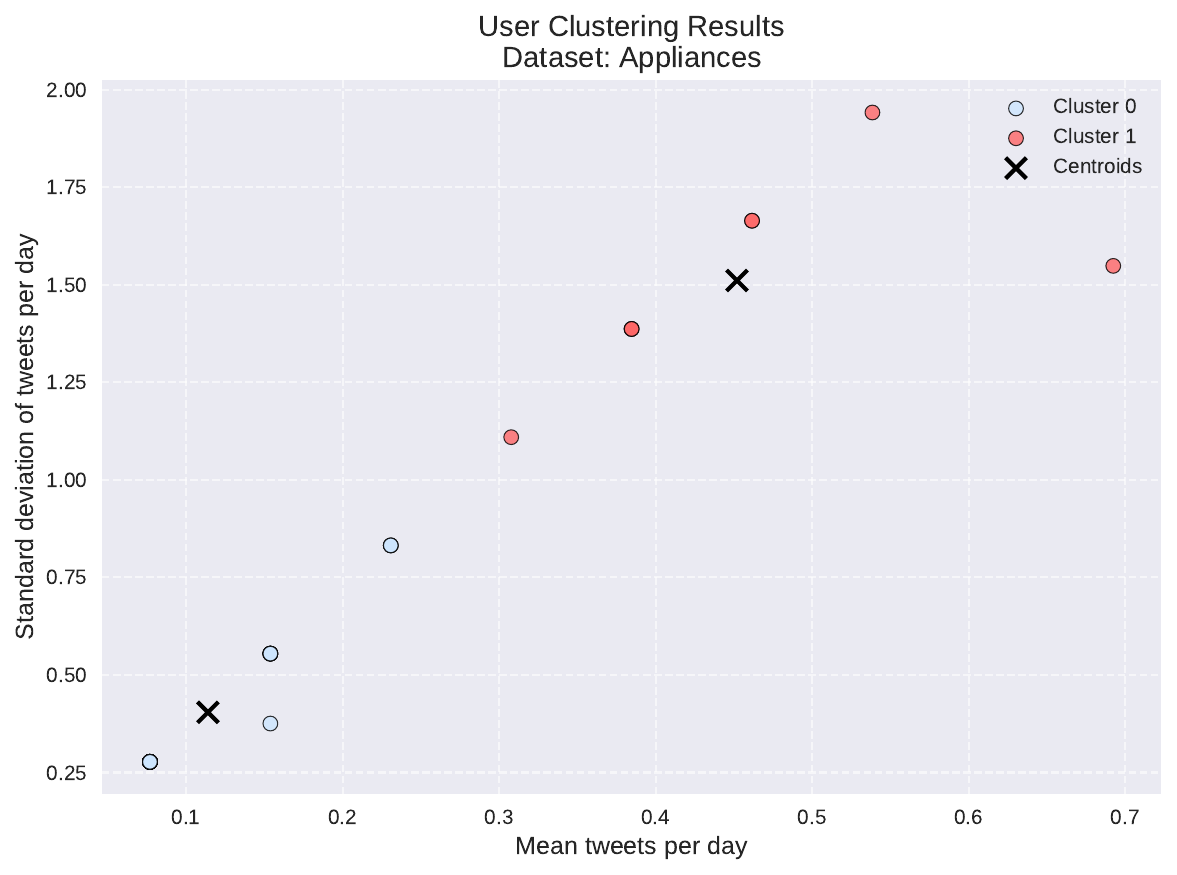}
    \caption{Appliances\_Mid-tail\_Split}
  \end{subfigure}
  \hfill
  \begin{subfigure}[b]{0.41\textwidth}
    \includegraphics[width=\textwidth]{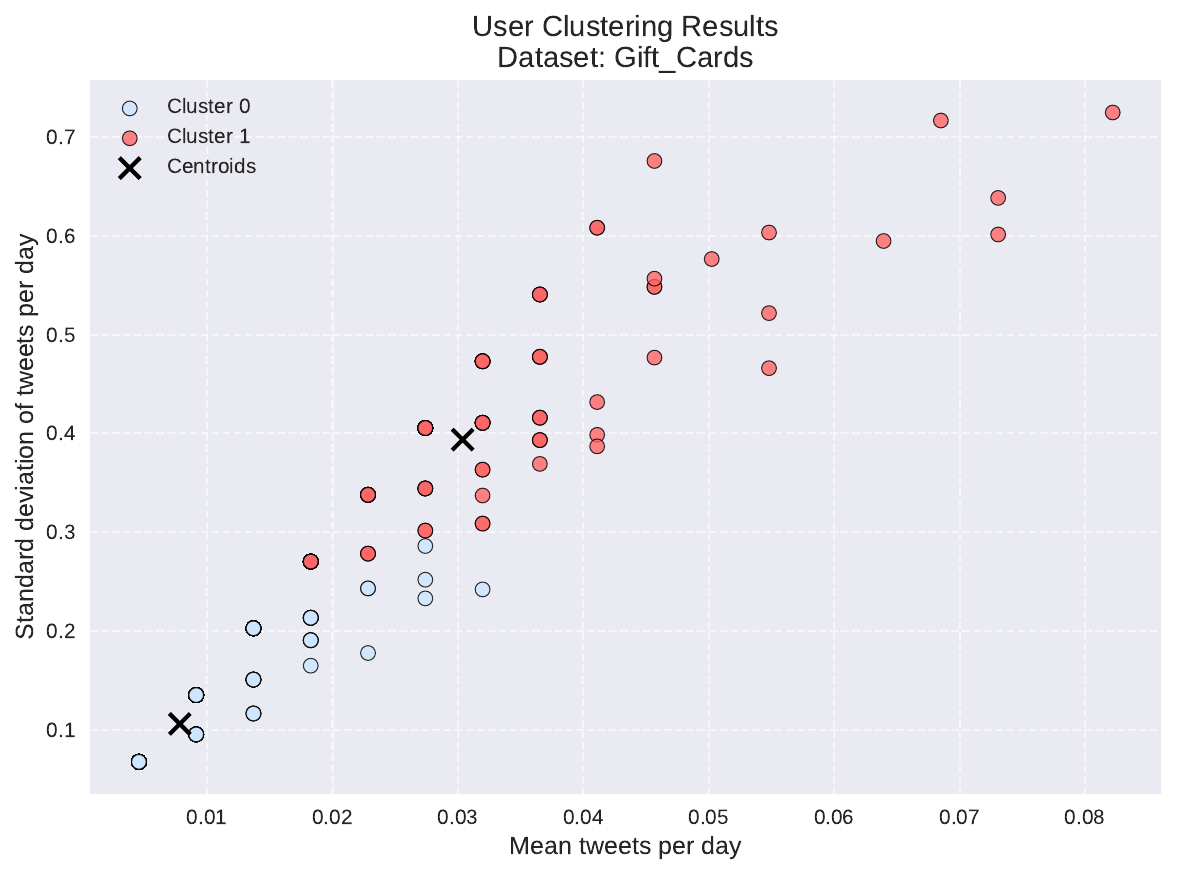}
    \caption{Gift\_Cards\_Mid-tail\_Split}
  \end{subfigure}
  \vskip\baselineskip 
  \begin{subfigure}[b]{0.41\textwidth}
    \includegraphics[width=\textwidth]{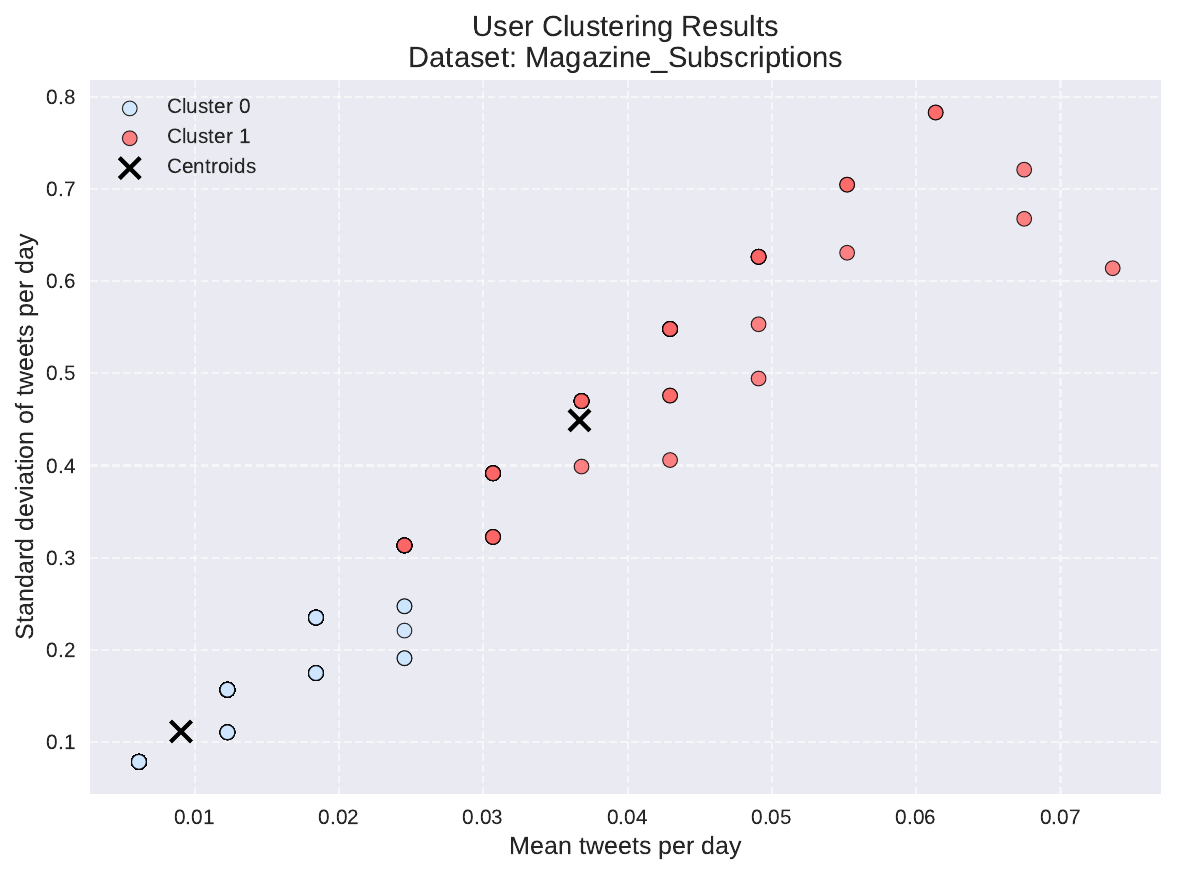}
    \caption{Magazine\_Subscriptions\_Mid-tail\_Split}
  \end{subfigure}
  \caption{Non-Data Sparse User Division. This section discusses users who are sparse in time rather than in data. The data points in the upper right corner indicate users with abundant but uneven data. The red dots in the figure are defined as mid-tail users.}
  \label{fig:nonsparse_division}
\end{figure*}

\noindent\textbf{Mid-tail User Identification.}  
Figure~\ref{fig:nonsparse_division} further divides non-sparse users who exhibit temporal sparsity. We compute the number of reviews per user per day and apply K-means clustering using statistical indicators such as mean, standard deviation, minimum, and maximum review counts. The top-right region in the figure represents highly active users with fluctuating review frequencies, while the top-left region corresponds to similarly active users with more stable review patterns. The bottom-right region consists of less active users who occasionally review in bursts, and the bottom-left region contains users with consistently low review activity. Based on these distributions, we define mid-tail users as those in the top-right and bottom-right groups.  

\begin{figure*}[htbp]
  \centering
  \begin{subfigure}[b]{0.45\textwidth}
    \includegraphics[width=\textwidth]{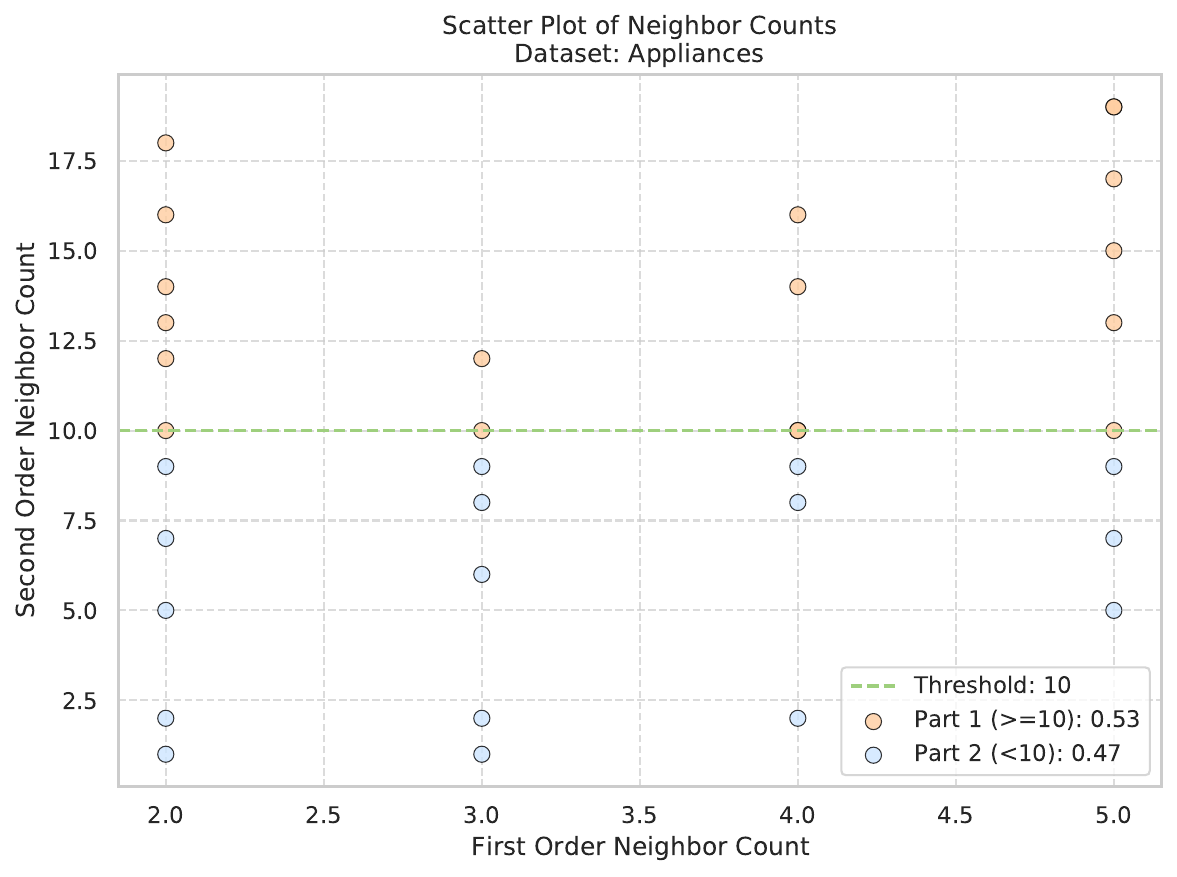}
    \caption{Appliances\_Long-tail\&Extreme\_Split}
  \end{subfigure}
  \hfill
  \begin{subfigure}[b]{0.45\textwidth}
    \includegraphics[width=\textwidth]{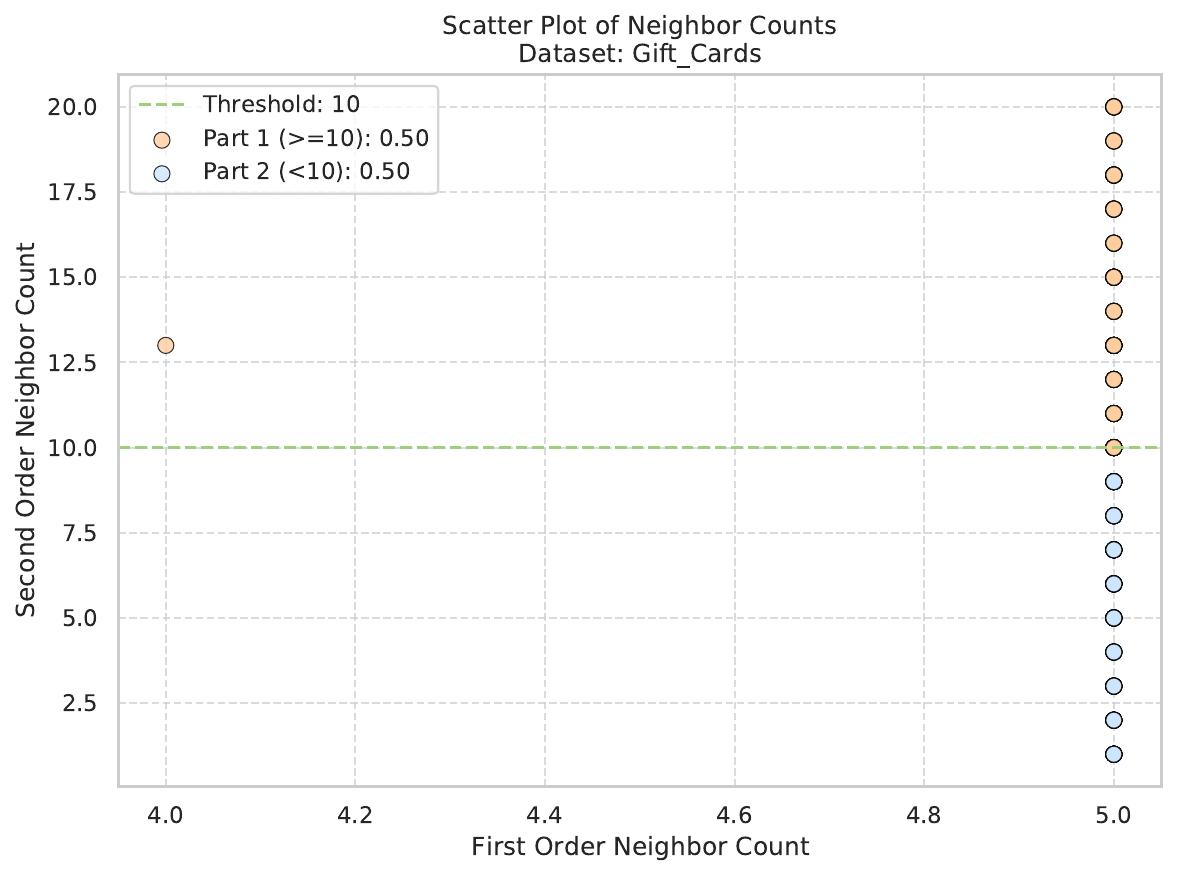}
    \caption{Gift\_Cards\_Long-tail\&Extreme\_Split}
  \end{subfigure}
  \vskip\baselineskip 
  \begin{subfigure}[b]{0.45\textwidth}
    \centering
    \includegraphics[width=\textwidth]{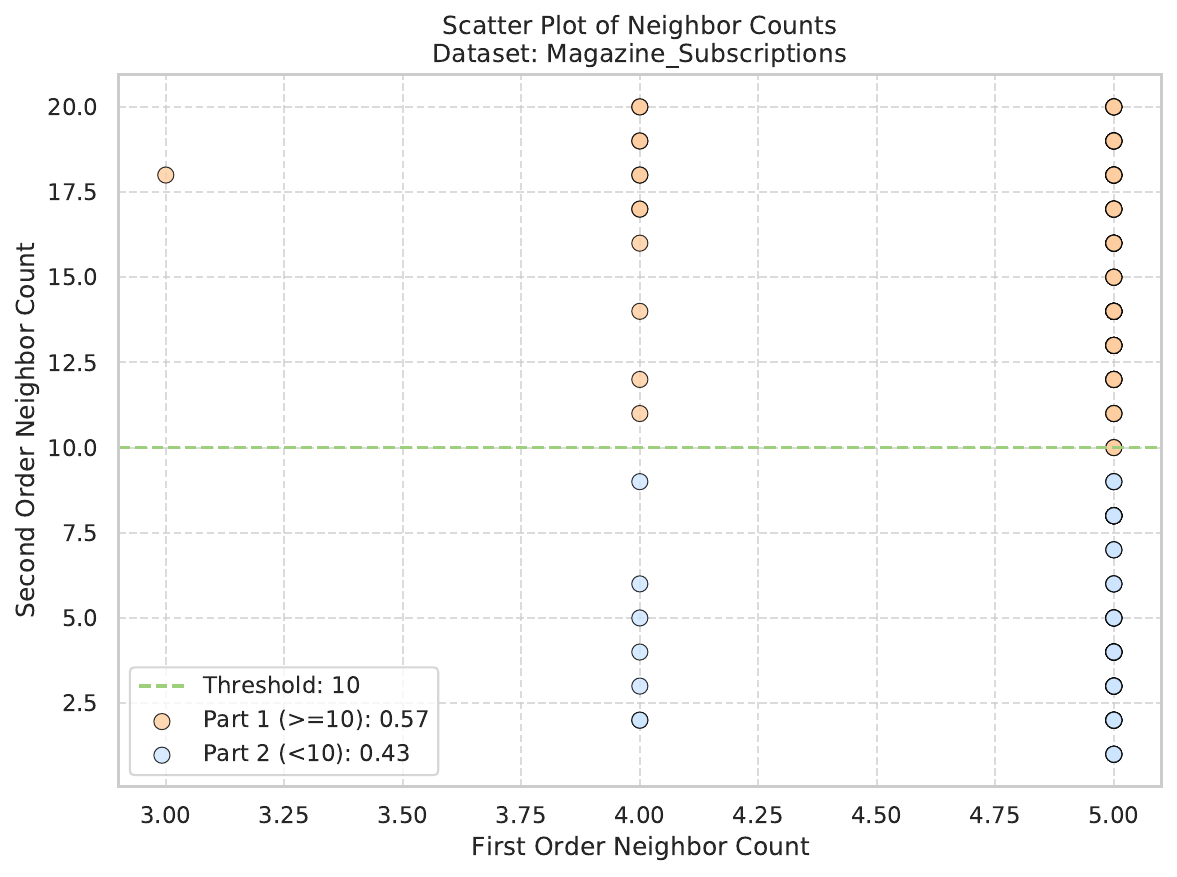}
    \caption{Magazine\_Subscriptions\_Long-tail\&Extreme\_Split}
  \end{subfigure}
  \caption{Data-Sparse User Division and Corresponding Proportions. The orange points exhibit abundant second-order homogeneous relationships and are defined as long-tail users, while the blue points have sparse second-order homogeneous relationships and are defined as extreme cases.}
  \label{fig:sparse_division}
\end{figure*}

\noindent\textbf{Long-tail and Extreme User Categorization.}  
Figure~\ref{fig:sparse_division} illustrates the distinction between long-tail and extreme users based on second-order neighborhood connectivity. Sparse users with limited self-data but sufficient second-order neighbors are classified as long-tail users, as their profiles can be supplemented with synthesized data. Conversely, users with both limited self-data and few second-order neighbors are categorized as extreme users, where direct data synthesis is required.  

\begin{figure*}[htbp]
  \centering
  \begin{subfigure}[b]{0.37\textwidth} 
    \centering
    \includegraphics[width=\textwidth]{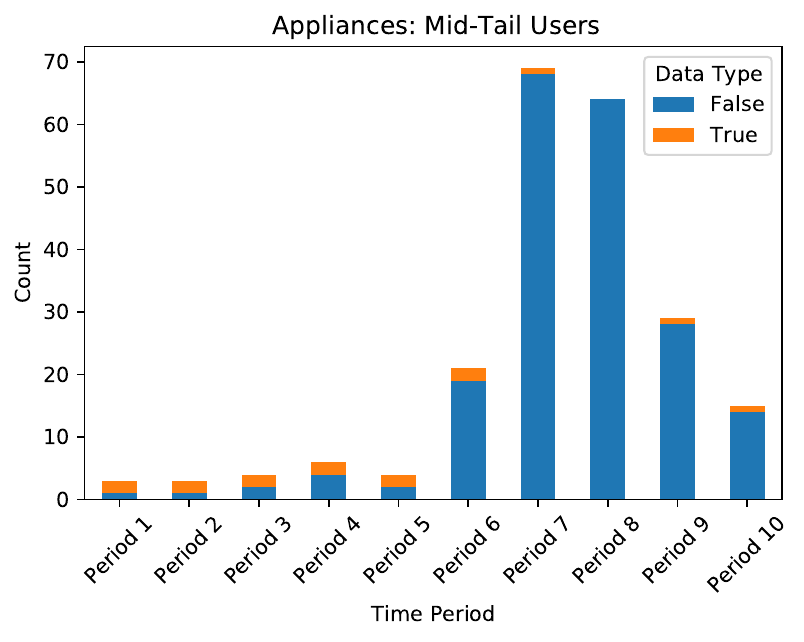}
    \caption{Appliances\_Mid-Tail}
  \end{subfigure}
  \hfill
  \begin{subfigure}[b]{0.37\textwidth}
    \centering
    \includegraphics[width=\textwidth]{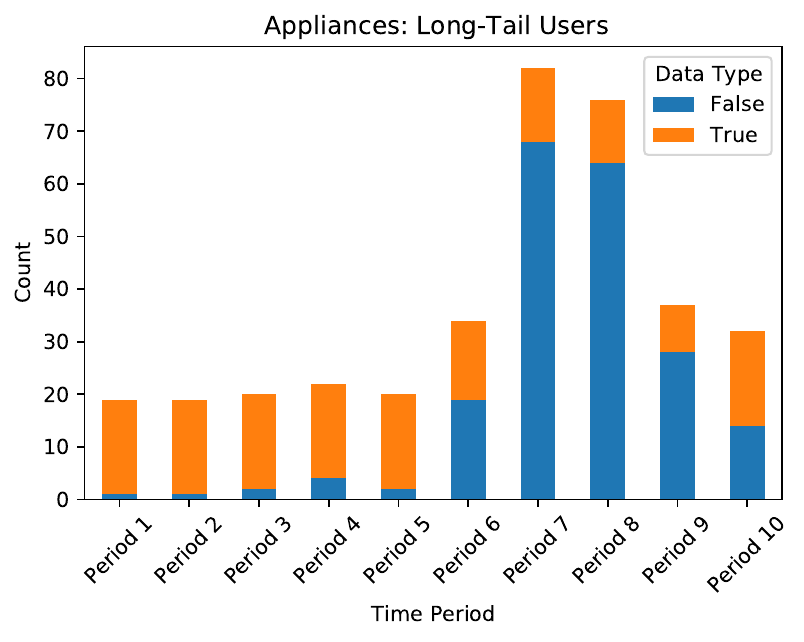}
    \caption{Appliances\_Long-Tail}
  \end{subfigure}

  \vspace{0.8em} 

  \begin{subfigure}[b]{0.37\textwidth}
    \centering
    \includegraphics[width=\textwidth]{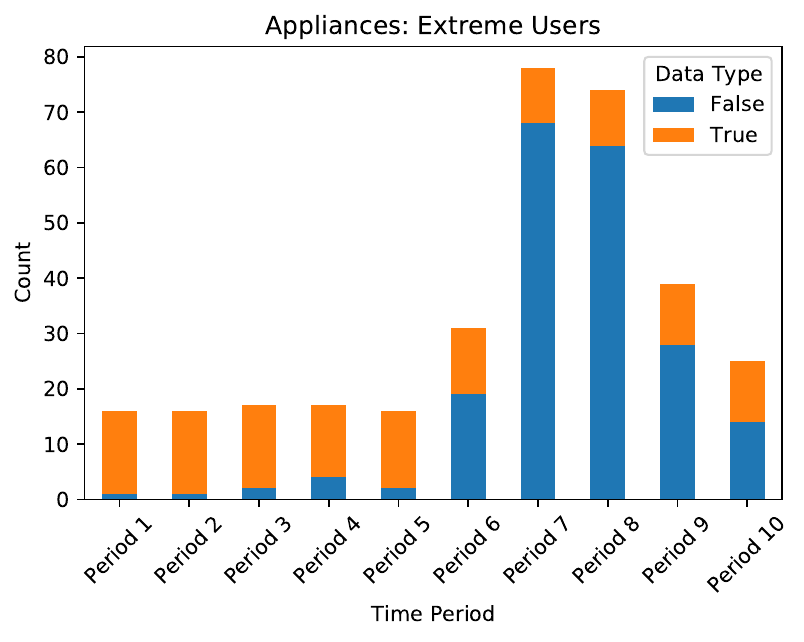}
    \caption{Appliances\_Extreme}
  \end{subfigure}
  \hfill
  \begin{subfigure}[b]{0.37\textwidth}
    \centering
    \includegraphics[width=\textwidth]{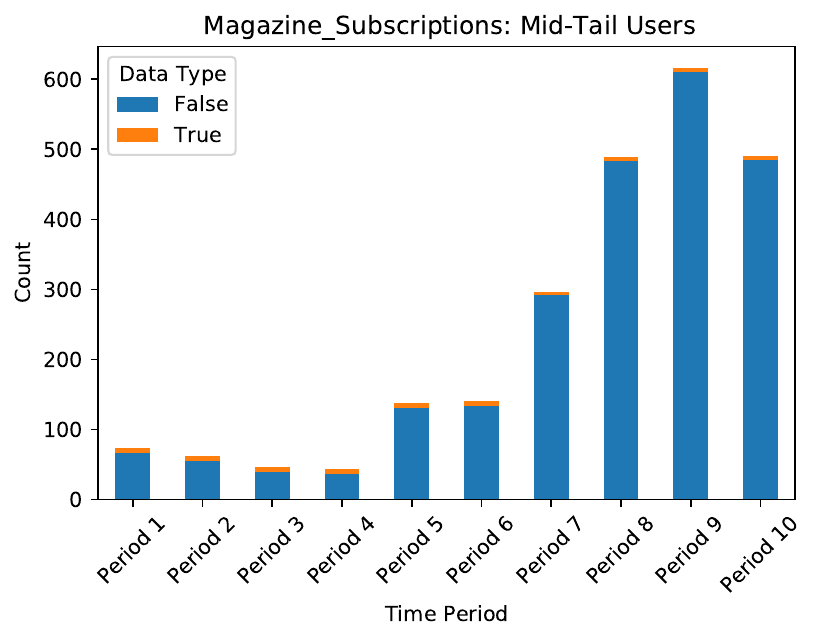}
    \caption{Magazine\_Subscriptions\_Mid-Tail}
  \end{subfigure}

  \vspace{0.8em}

  \begin{subfigure}[b]{0.37\textwidth}
    \centering
    \includegraphics[width=\textwidth]{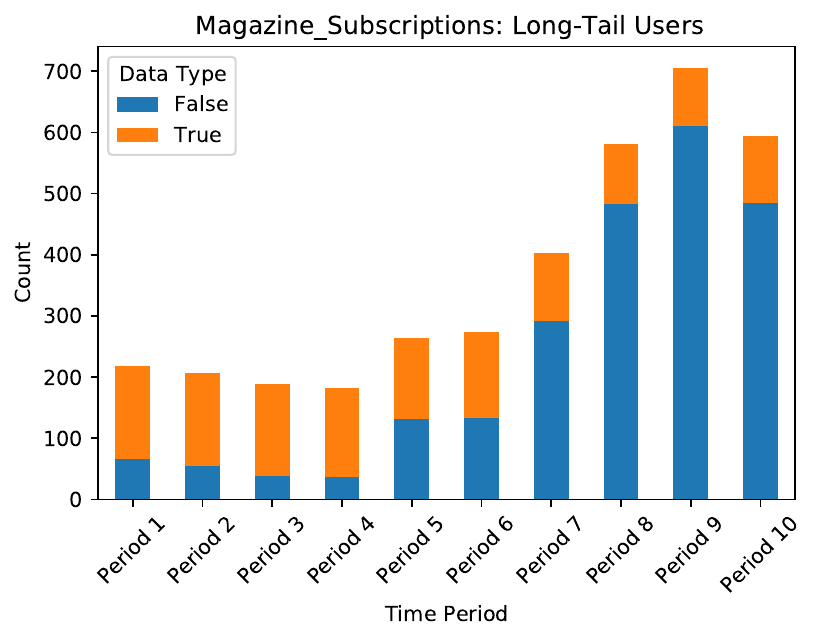}
    \caption{Magazine\_Subscriptions\_Long-Tail}
  \end{subfigure}
  \hfill
  \begin{subfigure}[b]{0.37\textwidth}
    \centering
    \includegraphics[width=\textwidth]{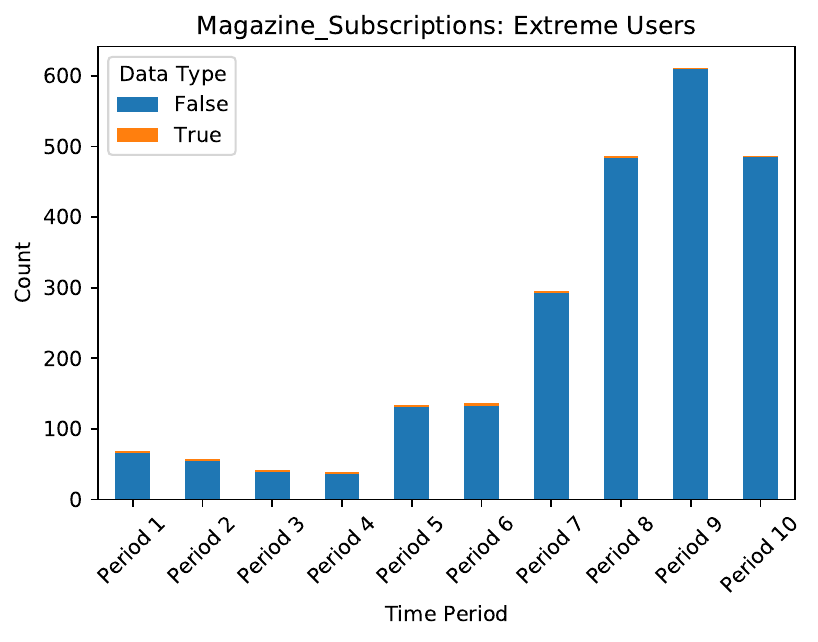}
    \caption{Magazine\_Subscriptions\_Extreme}
  \end{subfigure}

  \vspace{0.8em}

  \begin{subfigure}[b]{0.37\textwidth}
    \centering
    \includegraphics[width=\textwidth]{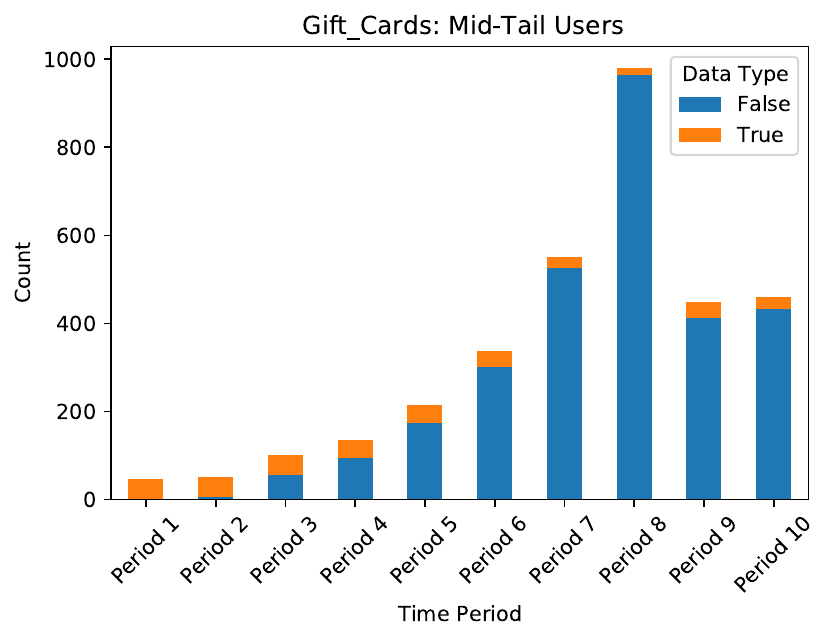}
    \caption{Gift\_Cards\_Mid-Tail}
  \end{subfigure}
  \hfill
  \begin{subfigure}[b]{0.37\textwidth}
    \centering
    \includegraphics[width=\textwidth]{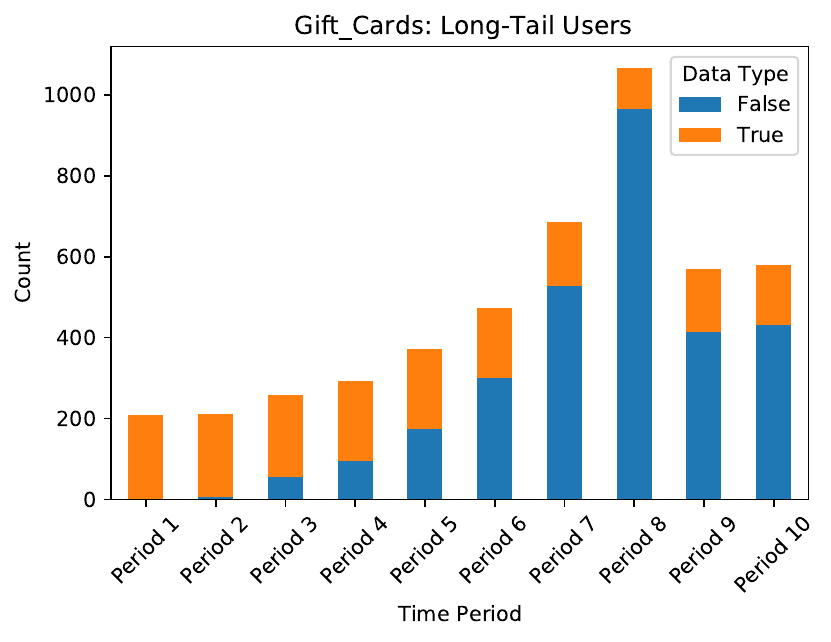}
    \caption{Gift\_Cards\_Long-Tail}
  \end{subfigure}

  \caption{Distribution of interpolation positions along the timeline corresponding to different sparse categories across datasets.}
  \label{fig:interpolated_distribution}
\end{figure*}

\noindent\textbf{Temporal Distribution of Interpolated Data.}  
In data interpolation, selecting insertion positions is crucial for synthesizing meaningful data. Figure~\ref{fig:interpolated_distribution} presents the temporal distribution of interpolated data across 10 time intervals.  

\section{Prompt Templates}
\label{sec:prompt}

\begin{figure*}[htbp]
\centering
\includegraphics[width=0.65\textwidth]{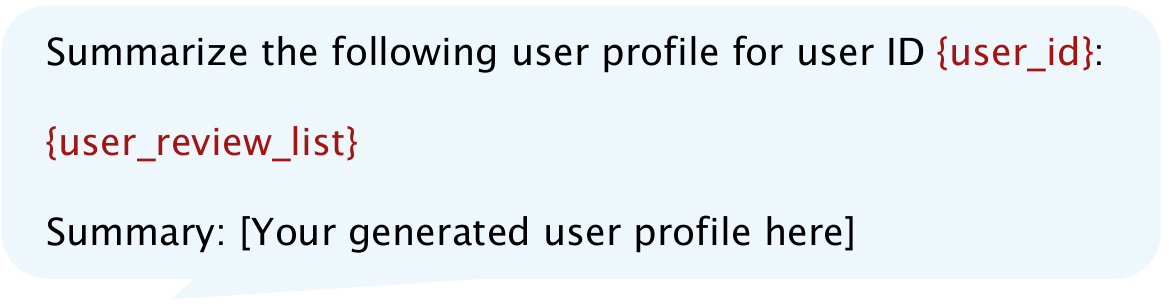}
\caption{The prompt used for generating user profiles in the mid-tail and extreme scenarios, defined as $\mathrm{P}_{um}$ and $\mathrm{P}_{ue}$ in the paper, respectively, takes as input the selected reviews of the user.}
\label{fig:P_ue}
\end{figure*}

\begin{figure*}[htbp]
\centering
\includegraphics[width=0.75\textwidth]{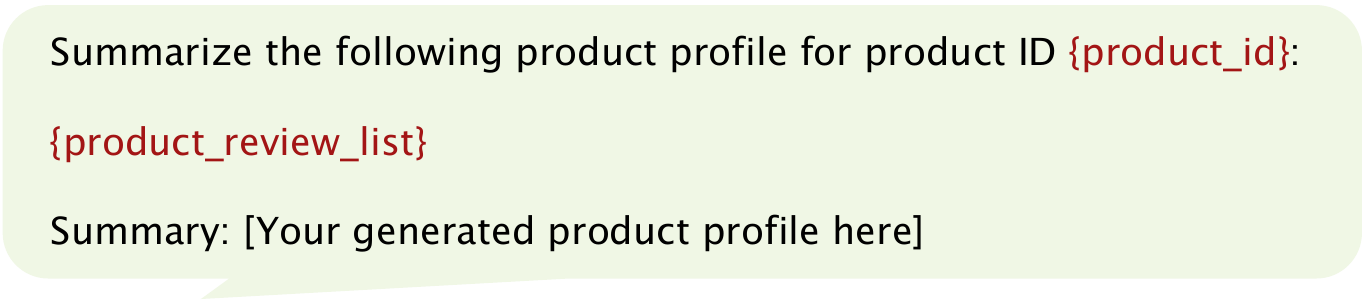}
\caption{The prompt used for generating product profiles in the mid-tail, long-tail, and extreme scenarios, defined as $\mathrm{P}_{pm}$, $\mathrm{P}_{pl}$, and $\mathrm{P}_{pe}$ in the paper, respectively, takes as input the selected reviews of the product.}
\label{fig:P_pe}
\end{figure*}

\begin{figure*}[htbp]
\centering
\includegraphics[width=0.9\textwidth]{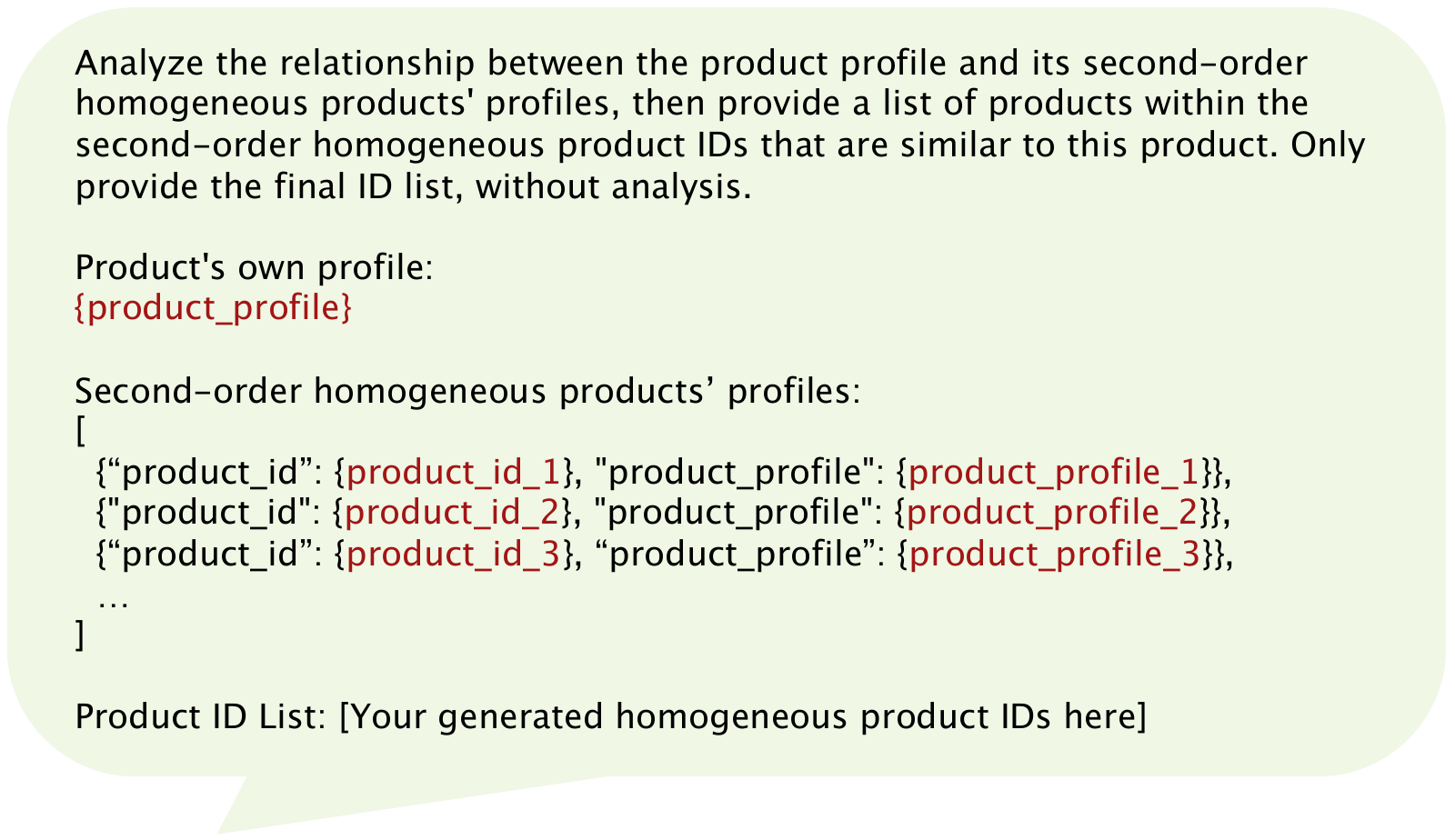}
\caption{The prompt used for selecting second-order homogeneous products in the mid-tail and long-tail scenarios, defined as $\mathrm{P}_{so}$ in the paper, takes as input the profile of the product itself along with the profile of the second-order homogeneous products.}
\label{fig:P_so}
\end{figure*}

\begin{figure*}[htbp]
\centering
\includegraphics[width=0.8\textwidth]{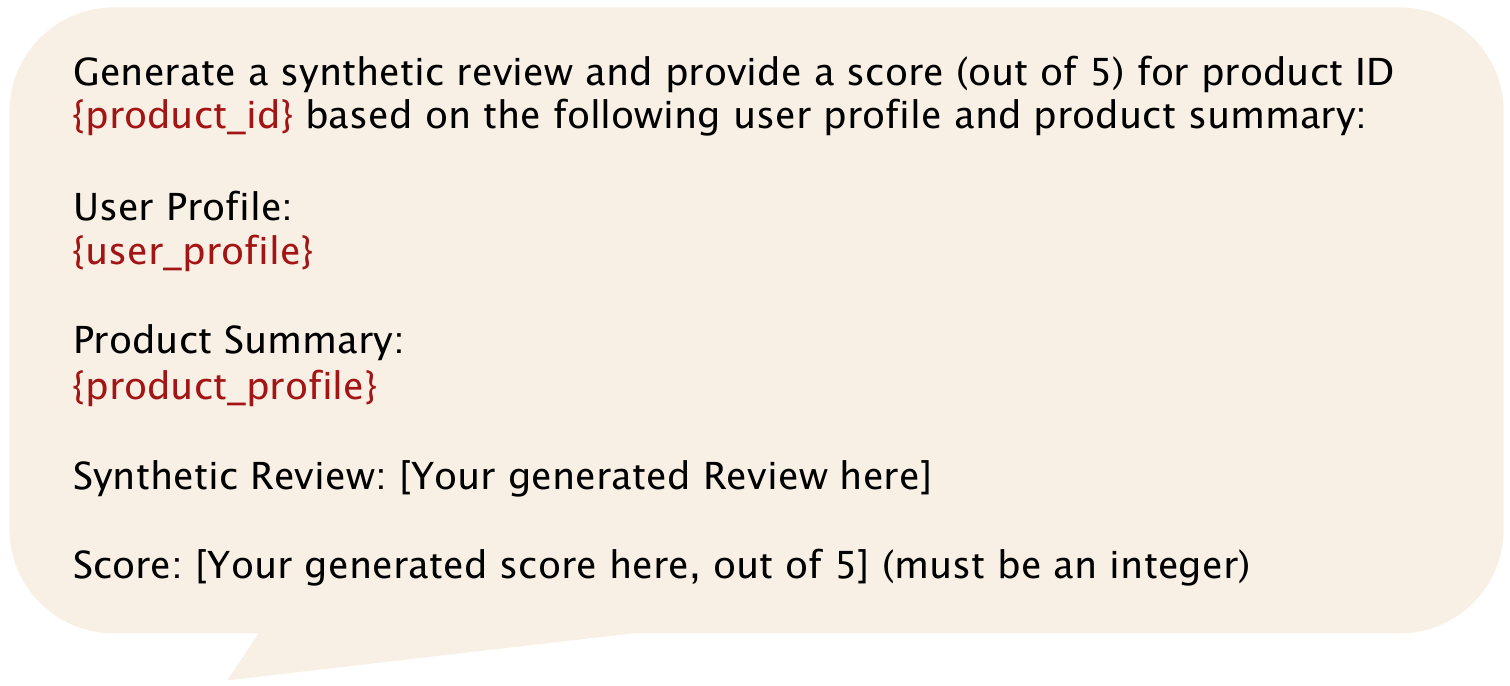}
\caption{The prompt used for inducing data for mid-tail, long-tail, and extreme user scenarios, defined as $\mathrm{P}_{sd}$ in the paper, takes as input the user profile and the product profile.}
\label{fig:P_ds}
\end{figure*}

\begin{figure*}[htbp]
\centering
\includegraphics[width=0.9\textwidth]{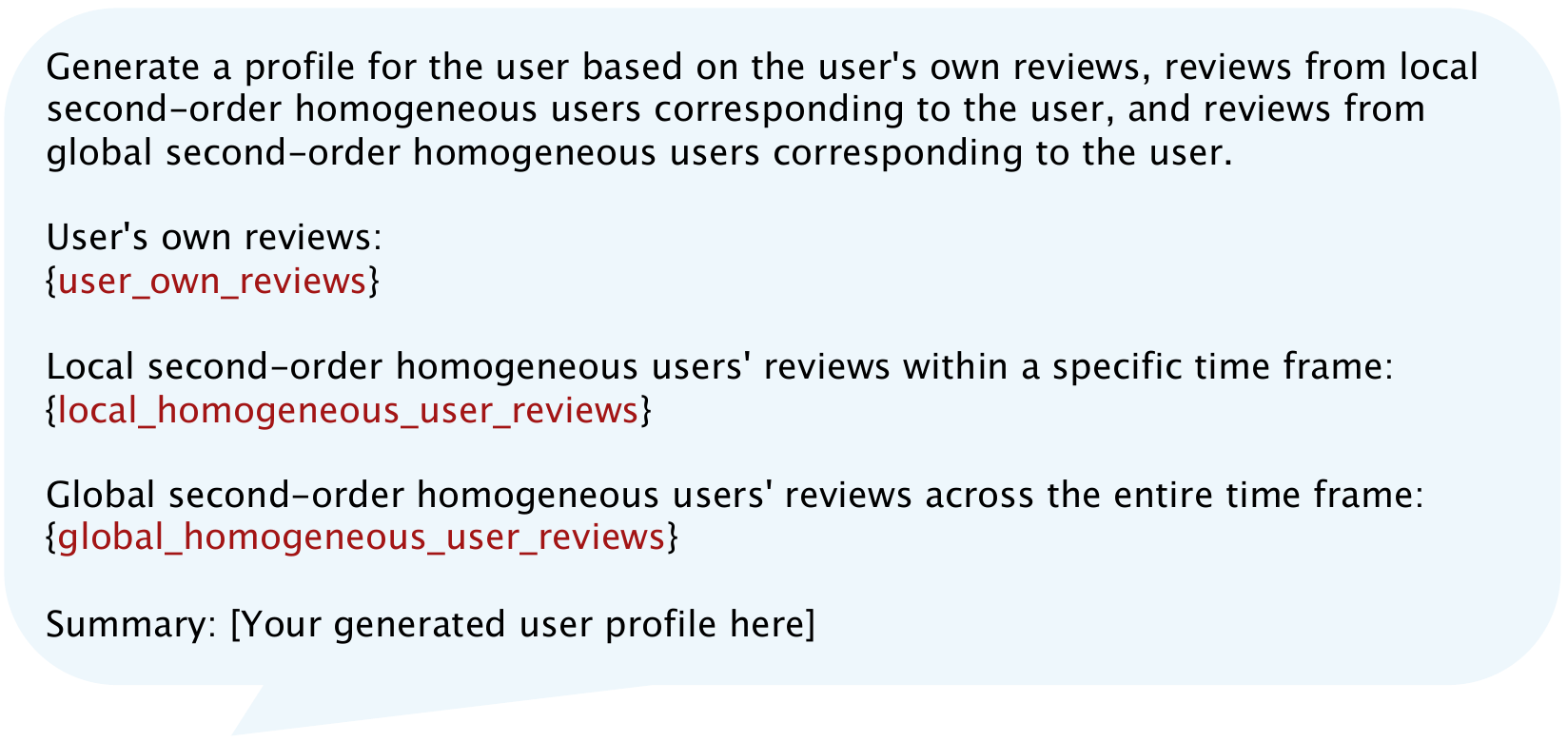}
\caption{The prompt used for understanding the local and global user relationships in the long-tail scenario, defined as $\mathrm{P}_{ul}$ in the paper, takes as input the user's own reviews, the reviews of locally second-order homogeneous users, and the reviews of globally second-order homogeneous users.}
\label{fig:P_ul}
\end{figure*}

We use OpenAI's gpt-3.5-turbo\footnote{\url{https://platform.openai.com/docs/api-reference/models}} for data synthesis, employing specific prompts tailored for different user categories. In the mid-tail scenario, $\mathrm{P}_{um}$ (Figure~\ref{fig:P_ue}) is used for user profile generation, $\mathrm{P}_{pm}$ (Figure~\ref{fig:P_pe}) for product profiles, $\mathrm{P}_{so}$ (Figure~\ref{fig:P_so}) for selecting second-order products, and $\mathrm{P}_{sd}$ (Figure~\ref{fig:P_ds}) for data synthesis. In the long-tail scenario, $\mathrm{P}_{ul}$ (Figure~\ref{fig:P_ul}) is used for generating user profiles, while $\mathrm{P}_{pl}=\mathrm{P}_{pm}$ is applied to product profiles. The prompts $\mathrm{P}_{so}$ and $\mathrm{P}_{sd}$ remain the same as in the mid-tail setting. For extreme users, $\mathrm{P}_{ue}=\mathrm{P}_{um}$ is used for user profiles, $\mathrm{P}_{pe}=\mathrm{P}_{pm}$ for product profiles, and $\mathrm{P}_{sd}$ follows the same prompt as in mid-tail users.  

\section{Synthetic-Data Proportion Experiments}

We conduct a series of controlled experiments on three datasets (\textit{Appliances}, \textit{Gift\_Cards}, \textit{Magazine\_Subscriptions}) using the DC-DGNN framework, in which synthetic data is inserted at six stages: 0\% (NoSynth), Front 20\%, Front 40\%, Front 60\%, Front 80\%, and 100\% (FullSynth).

\paragraph{Appliances (approximately 100 users)}  
In this extremely sparse setting, adding a small amount of synthetic data (Front 20\%) yields negligible changes in Accuracy or F1 and may introduce minor fluctuations due to noise. As the synthetic data ratio increases most metrics improve but Accuracy at Front 80\% slightly dips, which indicates that moderate augmentation often outperforms both no augmentation and full augmentation in very small datasets (see Table \ref{tab:appliances_synth}).

\paragraph{Gift\_Cards and Magazine\_Subscriptions (larger scale)}  
Both datasets show a generally monotonic increase in Accuracy and F1 as the synthetic data ratio increases, which suggests that when ample real data is available the negative impact of noise is reduced and the model benefits consistently from additional synthetic samples (see Tables \ref{tab:giftcards_synth} and \ref{tab:magazine_synth}).

\noindent Based on these results, we conclude that for very small and sparse datasets such as Appliances with only approximately 100 users moderate synthetic data ratios (for example 40 to 60\%) achieve the best balance between performance improvement and noise. Therefore, the synthetic data ratio and insertion stage must be calibrated on a per dataset basis.

\begin{table*}[htbp]
    \centering

    \begin{tabular}{lccccc}
        \toprule
        \textbf{Dataset} & 
        \textbf{Total R} & 
        \textbf{U10 R} & 
        \textbf{U10 R proportion} & 
        \textbf{U5 R} & 
        \textbf{U5 R proportion} \\ 
        \midrule
        Magazine\_Subscriptions & 2330 & 1178 & 0.506 & 764 & 0.328 \\
        Appliances             & 203  & 87   & 0.429 & 116 & 0.571 \\
        Gift\_Cards            & 2966 & 1502 & 0.506 & 1044 & 0.352 \\
        \bottomrule
    \end{tabular}
       \caption{%
    Statistical analysis of the ratio of user-associated reviews to the total review count across various hierarchical levels. 
    U10 R refers to the number of reviews associated with users with ten or fewer reviews. 
    U5 R refers to the number of reviews associated with users with five or fewer reviews.
    }  
    \label{tab:dif_user_statistic}
\end{table*}

\begin{table*}[htbp]
    \centering
    \resizebox{1.8\columnwidth}{!}{
    \begin{tabular}{lccccccc}
        \toprule
        \textbf{Variation} & \textbf{Accuracy} & \textbf{Precision} & \textbf{Recall} & \textbf{F1} & \textbf{MSE} & \textbf{RMSE} & \textbf{MAE} \\ 
        \midrule
        Appliances\_0\%\_NoSynth  & 0.7143 & 0.2381 & 0.3333 & 0.2778 & 1.0476 & 1.0235 & 0.4762 \\
        Appliances\_Front20\%     & 0.7143 & 0.2381 & 0.3333 & 0.2778 & 1.0476 & 1.0235 & 0.4762 \\
        Appliances\_Front40\%     & 0.7619 & 0.4250 & 0.4778 & 0.4492 & 0.5238 & 0.7237 & 0.3333 \\
        Appliances\_Front60\%     & 0.7619 & 0.4556 & 0.5389 & 0.4889 & 0.7619 & 0.8729 & 0.3810 \\
        Appliances\_Front80\%     & 0.7143 & 0.4000 & 0.4556 & 0.4222 & 0.5714 & 0.7559 & 0.3810 \\
        Appliances\_FullSynth     & 0.8571 & 0.5441 & 0.5833 & 0.5625 & 0.6667 & 0.8165 & 0.2857 \\
        \bottomrule
    \end{tabular}
    }
    \caption{Results on the Appliances dataset under different synthetic data proportions.}
    \label{tab:appliances_synth}
\end{table*}

\begin{table*}[htbp]
    \centering
    \resizebox{1.8\columnwidth}{!}{
    \begin{tabular}{lccccccc}
        \toprule
        \textbf{Variation} & \textbf{Accuracy} & \textbf{Precision} & \textbf{Recall} & \textbf{F1} & \textbf{MSE} & \textbf{RMSE} & \textbf{MAE} \\ 
        \midrule
        GiftCards\_0\%\_NoSynth  & 0.8754 & 0.2189 & 0.2500 & 0.2334 & 0.3064 & 0.5535 & 0.1717 \\
        GiftCards\_Front20\%     & 0.8721 & 0.1780 & 0.1992 & 0.1880 & 0.1751 & 0.4184 & 0.1414 \\
        GiftCards\_Front40\%     & 0.8721 & 0.4732 & 0.2897 & 0.3064 & 0.1481 & 0.3849 & 0.1347 \\
        GiftCards\_Front60\%     & 0.8923 & 0.3787 & 0.3667 & 0.3706 & 0.1279 & 0.3577 & 0.1145 \\
        GiftCards\_Front80\%     & 0.8990 & 0.7241 & 0.5086 & 0.5030 & 0.1212 & 0.3482 & 0.1077 \\
        GiftCards\_FullSynth     & 0.8956 & 0.4384 & 0.5000 & 0.4671 & 0.1145 & 0.3383 & 0.1077 \\
        \bottomrule
    \end{tabular}
    }
    \caption{Results on the Gift\_Cards dataset under different synthetic data proportions.}
    \label{tab:giftcards_synth}
\end{table*}

\begin{table*}[htbp]
    \centering
    \resizebox{1.8\columnwidth}{!}{
    \begin{tabular}{lccccccc}
        \toprule
        \textbf{Variation} & \textbf{Accuracy} & \textbf{Precision} & \textbf{Recall} & \textbf{F1} & \textbf{MSE} & \textbf{RMSE} & \textbf{MAE} \\ 
        \midrule
        Magazine\_0\%\_NoSynth   & 0.7554 & 0.4290 & 0.4107 & 0.4016 & 0.7768 & 0.8814 & 0.3820 \\
        Magazine\_Front20\%      & 0.7639 & 0.5851 & 0.4753 & 0.4830 & 0.5193 & 0.7206 & 0.3133 \\
        Magazine\_Front40\%      & 0.7768 & 0.4126 & 0.4794 & 0.4370 & 0.5064 & 0.7116 & 0.3004 \\
        Magazine\_Front60\%      & 0.8155 & 0.7670 & 0.6302 & 0.6210 & 0.3133 & 0.5597 & 0.2189 \\
        Magazine\_Front80\%      & 0.8155 & 0.5576 & 0.5495 & 0.4991 & 0.3906 & 0.6249 & 0.2361 \\
        Magazine\_FullSynth      & 0.7983 & 0.6879 & 0.5853 & 0.5385 & 0.4206 & 0.6485 & 0.2575 \\
        \bottomrule
    \end{tabular}
    }
    \caption{Results on the Magazine\_Subscriptions dataset under different synthetic data proportions.}
    \label{tab:magazine_synth}
\end{table*}

\begin{figure*}[htbp]
\centering
\includegraphics[width=0.8\textwidth]{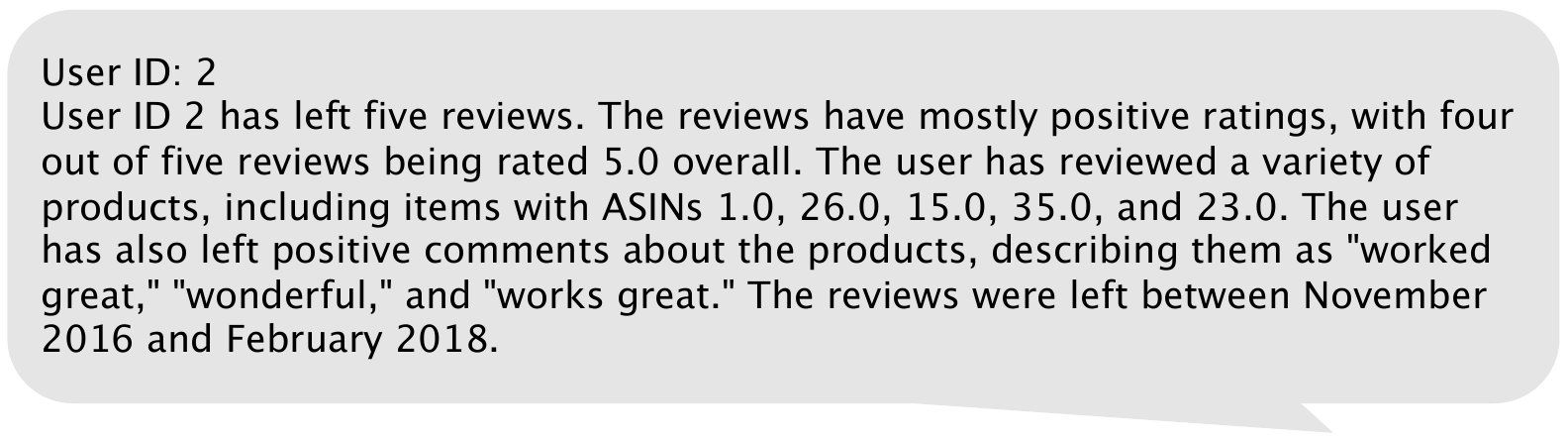}
\caption{Example of user profiles generated by GPT.}
\label{fig:U_exp}
\end{figure*}

\begin{figure*}[htbp]
\centering
\includegraphics[width=0.8\textwidth]{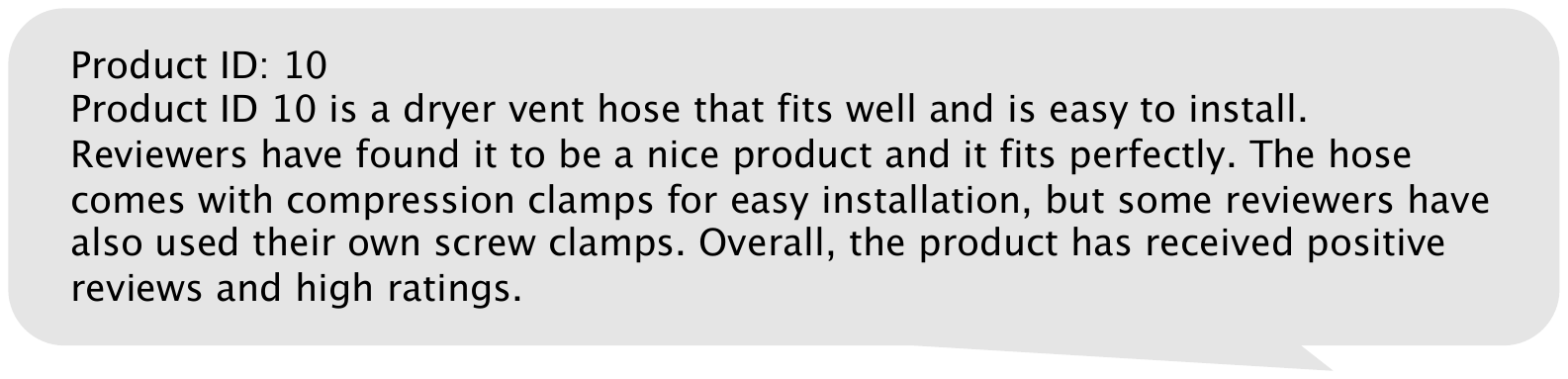}
\caption{Example of product profiles generated by GPT.}
\label{fig:P_exp}
\end{figure*}

\begin{figure*}[htbp]
\centering
\includegraphics[width=0.9\textwidth]{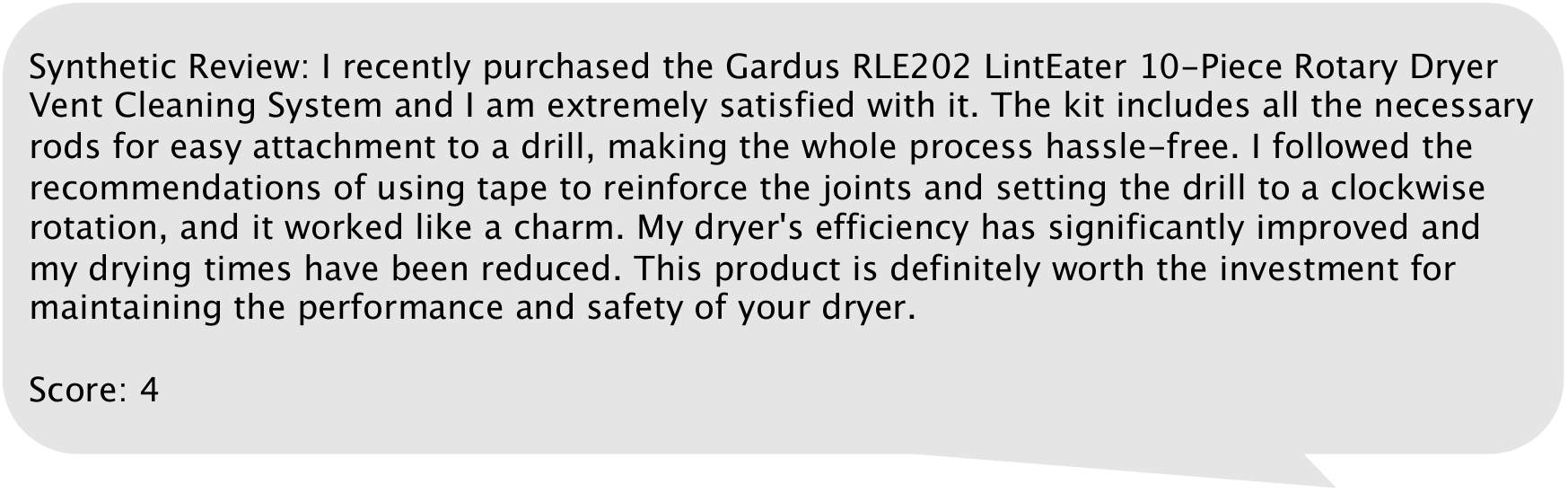}
\caption{Example of synthesized data demonstrating positive sentiment generated by GPT.}
\label{fig:sd_exp1}
\end{figure*}

\begin{figure*}[htbp]
\centering
\includegraphics[width=0.9\textwidth]{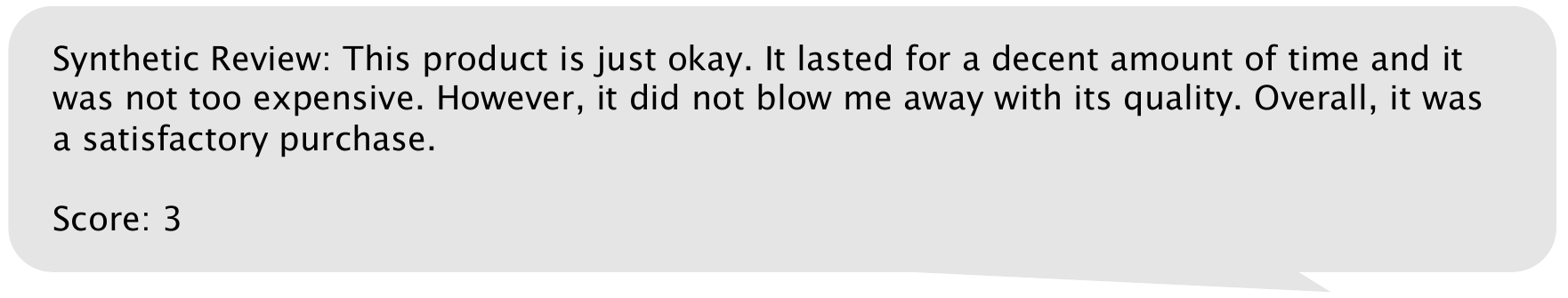}
\caption{Example of synthesized data demonstrating neutral sentiment generated by GPT.}
\label{fig:sd_exp2}
\end{figure*}


\end{document}